\documentclass[10pt,twocolumn,letterpaper]{article}
\usepackage{cvpr}      % To produce the REVIEW version
\usepackage[ruled,linesnumbered,noline]{algorithm2e}
\SetAlgoNlRelativeSize{0}     % 行号字号与正文相同
\SetNlSty{}{}{：}              % 行号格式：1: 2: 3:
\SetAlgoNlRelativeSize{0}    % 如果希望略小可以改成 -1
\SetNlSkip{0.4em}             % 行号与文字之间的水平间距

\usepackage{multirow} % 用于垂直居中
\usepackage{booktabs}
\definecolor{cvprblue}{rgb}{0.21,0.49,0.74}
\usepackage[pagebackref,breaklinks,colorlinks,allcolors=cvprblue]{hyperref}
\usepackage[american]{babel}
\usepackage{microtype}
\usepackage{multirow}
\usepackage{graphicx}
\usepackage{float}
\usepackage{dblfloatfix}  % 修复双栏浮动体问题
\usepackage{placeins}     % 支持 \FloatBarrier
\usepackage{algorithmic}
\usepackage{bm}
\usepackage{tabularx}

\def\paperID{}
\def\confName{CVPR}
\def\confYear{2026}

\title{Vasicek Graph Diffusion Model}
\author{Junyi Tang, Chang Liu, Kun Zhan$^\star$\\
School of AI, Lanzhou University}
\begin{document}
\maketitle
\begin{abstract}
Diffusion-based graph generation models progressively transform Gaussian noise into structured graphs through stochastic dynamics. We observe that the effectiveness of this transformation is fundamentally determined by the evolution of the diffusion trajectory. Existing graph diffusion models often suffer from excessive uncertainty during diffusion and delayed growth of the signal-to-noise ratio, causing useful structural information to emerge only in the later stages of generation. To address this issue, we propose the Vasicek Graph Diffusion (VGD), a graph diffusion process derived from the Vasicek dynamics. The key insight of VGD is data-anchored mean dynamics with a closed-form solution, which analytically establishes a continuous transition from Gaussian noise to the target graph distribution. Unlike conventional graph diffusion processes that rely predominantly on stochastic evolution, VGD introduces a data-anchored drift that continuously steers the latent state toward the target distribution throughout diffusion. As a consequence, VGD maintains a smoothly evolving uncertainty while rapidly increasing the signal-to-noise ratio throughout diffusion, allowing informative graph structures to emerge much earlier during generation. Extensive experiments on multiple graph generation benchmarks demonstrate that VGD outperforms existing diffusion-based graph generative models in terms of generation quality and structural fidelity.
\end{abstract}

\section{Introduction}
Diffusion-based generative models have emerged as a powerful framework for modeling complex data distributions through stochastic differential equations (SDEs)~\cite{10.5555/3495724.3496298,song2021scorebased}.
% By progressively perturbing structured data into noise and learning to reverse this process, these models have achieved remarkable success in continuous domains such as drug discovery, circuit design, and social simulation.
% By progressively perturbing data into noise starting from a simple prior distribution and learning to reverse this process, 
% These models are capable of constructing highly intricate samples. 
This generative capability has led to breakthrough progress in modeling structured graph data, enabling remarkable success across challenging domains ranging from molecular design to social network simulation~\cite{jo2022score,vignac2022digress,jo2023graph,kong2023autoregressive}.

\begin{figure}[t]
    \centering
    % \hspace*{-8mm}  
    \includegraphics[width=1\columnwidth]{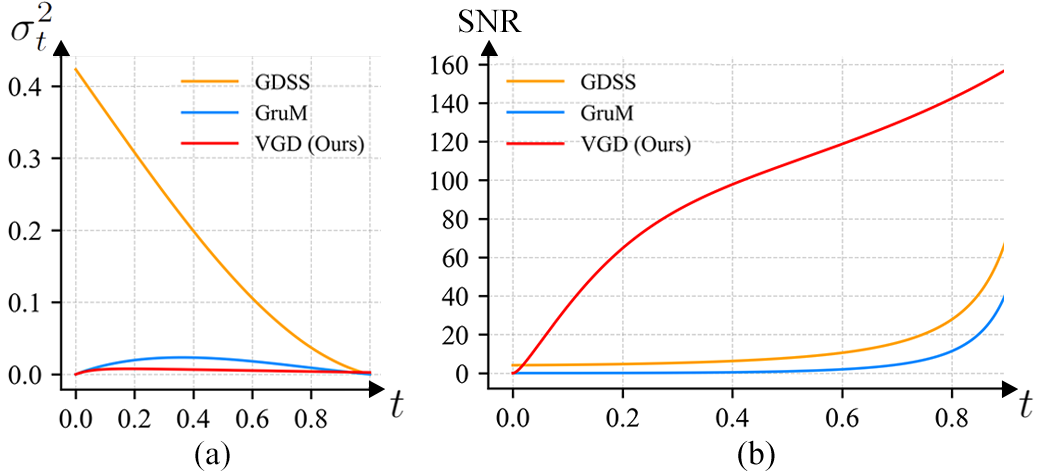}
\caption{Diffusion dynamics analysis of VGD compared with existing graph diffusion models.
(a) Conditional variance evolution from noise to data during the diffusion process. 
(b) Signal-to-noise ratio (SNR) evolution along the same noise-to-data trajectory.
All variance curves are analytically computed from the corresponding transition kernels.
% , where the conditional variance is ${\rm Var}[x_t|x_0,x_T]$ for endpoint-anchored diffusion processes and ${\rm Var}[x_t|x_T]$ for GDSS~\cite{jo2022score}. 
Compared with GDSS and GruM, VGD achieves a smoother uncertainty transition with consistently controlled variance, while enabling earlier and more progressive signal recovery throughout the diffusion trajectory.
These observations motivate the design of VGD, where the diffusion dynamics are explicitly constructed to provide a stable and informative transition from noise to structured graph data.
}

    \label{fig:var_GDSS}
\end{figure}

Recent extensions, including graph diffusion models like GDSS~\cite{jo2022score} and diffusion bridges such as GruM~\cite{jo2023graph}, have demonstrated the potential of stochastic dynamics for generating structured graph data with high fidelity. However, the effectiveness of graph diffusion is fundamentally determined by how the stochastic trajectory evolves from noise to data~\cite{10.5555/3495724.3496298,song2021scorebased,liu2024graph}. As shown in Figure~\ref{fig:var_GDSS}, existing approaches often exhibit undesirable diffusion dynamics during the transition from noise to data. 
Large diffusion variance introduces substantial uncertainty into intermediate states, making accurate denoising difficult, while a prolonged low signal-to-noise ratio (SNR) delays the emergence of informative structural signals throughout the trajectory. 
These limitations are particularly pronounced in graph generation, where complex dependencies among nodes and edges require efficient recovery of structural information from noisy states~\cite{YuICLR2025spp}. 

To address these challenges, we propose the Vasicek Graph Diffusion (VGD), a novel graph diffusion framework derived from the classical Vasicek stochastic process~\cite{VASICEK1977177}. 
Unlike conventional mean-reverting processes that converge toward a fixed equilibrium, VGD introduces a data-anchored stochastic dynamics, where the equilibrium point is defined by the target graph itself. 
This design naturally establishes an analytically tractable diffusion between Gaussian prior  and structured data, with closed-form mean and variance evolution throughout the trajectory.
% For graph diffusion models, maintaining a well-controlled bridge between noise and data without compromising training stability remains a persistent challenge~\cite{de2021diffusion,yang2023directional}. The core of the problem lies in the design of the drift and the variance schedule, which often results in turbulent paths that are difficult to learn. To address these challenges and achieve stable, analytically tractable dynamics, we propose the Vasicek Diffusion Model (VGD), a novel graph generation framework based on the Vasicek stochastic process~\cite{VASICEK1977177}. The Vasicek process is characterized by its mean-reverting property, which naturally guides the process toward a predefined target. We leverage this mathematical property by initializing the process with a noise prior and setting the target graph as the intrinsic reversion target. By anchoring the dynamics directly to the target data, VGD constructs a analytically tractable diffusion process from noise to complex graph structures.

The key insight of VGD lies in its data-anchored drift, which induces an exponentially weighted interpolation between the prior noise distribution and the target graph distribution. 
This analytically tractable dynamics establishes a direct and smooth transition toward the data manifold, avoiding prolonged diffusion trajectories dominated by stochastic uncertainty. 
As illustrated in Figure~\ref{fig:var_GDSS}, VGD achieves a diffusion trajectory by simultaneously maintaining a low and smoothly evolving variance while rapidly increasing the SNR from early diffusion stages. 
Such dynamics facilitate the early recovery of informative graph structures during generation, enabling more effective utilization of the sampling trajectory for refining graph topology. 
Consequently, VGD provides a stable and efficient diffusion paradigm for graph generation.

Our main contributions are summarized as follows:
\begin{itemize}[leftmargin=*]
    % Vasicek Diffusion Bridge (VGD):
\item We introduce a novel diffusion model, Vasicek Graph Diffusion (VGD). It defines closed-form mean dynamics that analytically bridges Gaussian noise and data distributions, enabling both theoretical tractability and efficient sampling. 
% , which can be viewed as a variance-stabilized form of the Ornstein–Uhlenbeck bridge. 
% The VGD formulation ensures smoother variance evolution, facilitating stable learning of intermediate states.
% 我们引入了一种源自Vasicek过程的新型扩散桥模型，该模型可以看作是Ornstein-Uhlenbeck桥的方差稳定形式。VGD模型确保了更平滑的方差演化，从而有助于中间状态的稳定学习。
    \item We reveal that excessive or rapidly changing variance in conventional diffusion models hinders denoising. To address this, VGD introduces a smoothly evolving variance schedule that maintains lower variance across time.%, resulting in more consistent and stable network learning.
    % Variance Reduction Regularization:
% To further constrain the variance dynamics, we design an additional regularization term. This mechanism mitigates variance oscillation during both training and sampling, leading to faster and more consistent convergence.
% 为了进一步约束方差动态变化，我们设计了一个额外的正则化项。该机制可以缓解训练和采样过程中的方差振荡，从而实现更快、更稳定的收敛。
%     \item Efficient and Reliable Graph Generation:
% By integrating Vasicek dynamics with variance reduction, our method achieves higher validity, novelty, and diversity in molecular graph generation while maintaining computational efficiency.
% 高效可靠的图生成：
% 通过将 Vasicek 动力学与方差缩减相结合，我们的方法在保持计算效率的同时，实现了分子图生成的更高有效性、新颖性和多样性。
    \item Through trajectory analysis, we show that VGD reduces noise-dominated wandering during early sampling and approaches the target modes earlier than the bridge baseline. 

% Extensive experiments on graph generation benchmarks demonstrate that VGD achieves superior performance in terms of structural fidelity and generative quality., outperforming prior SDE-based methods such as GDSS and GruM.
    % Theoretical and Empirical Validation:
% We provide analytical insight into the variance behavior of Vasicek bridges and conduct extensive experiments on molecular benchmarks, verifying the superiority of our approach over existing Ornstein–Uhlenbeck  stochastic process-based and score-driven graph diffusion models.
% 理论与实证验证：
% 我们对Vasicek桥的方差行为进行了分析，并在分子基准测试上进行了大量实验，验证了我们的方法优于现有的基于Ornstein–Uhlenbeck stochastic process和评分驱动的图扩散模型。
\end{itemize}

Experiments on molecular and general-purpose graph benchmarks further demonstrate improved structural fidelity and generative quality. 

\section{Related Work}
% \subsection{Diffusion Models}
Diffusion models and their continuous variants score-based generative models~\cite{song2019generative,song2020improved}, construct a forward noising process from the data distribution to a prior distribution, and learn the reverse denoising process to enable sample generation. In the discrete-time framework, DDPMs establish a stepwise noise diffusion process on a Markov chain, 
where the reverse process is parameterized by a learnable network that predicts the noise to reconstruct samples approximating the true data distribution~\cite{songdenoising,rombach2022high,nichol2021improved,austin2021structured,liu2024residual,nichol2022glide}. Score-based models, on the other hand, adopt a continuous-time stochastic differential equation (SDE) formulation and perform reverse sampling by estimating the score function. These approaches have demonstrated outstanding generative quality and strong theoretical foundations across domains such as image, audio, and graph generation~\cite{yoon2023score,lim2023score,kongdiffwave,jo2023graph}. %For tractable training and sampling, the forward noising process is engineered such that its target distribution has a minimal Kullback-Leibler (KL)~\cite{sohl2015deep} divergence from a simple and tractable prior. However, the mismatch between the forward process's target distribution and the reverse sampling prior degrades generation quality. To address this issue, VGD defines an analytically tractable noise-to-data process that aligns training and generation by initializing both the conditional training trajectories and the generative trajectories from the same prescribed Gaussian prior.
Graph generative models learn the joint distribution over nodes and edges, enabling the generation of novel graph structures that satisfy required structural constraints~\cite{de2018molgan,wang2025fdgen,chengraph}. Early methods often employed autoregressive or variational graph decoders to generate nodes and edges~\cite{simonovsky2018graphvae,jin2018junction,shi2020graphaf,ommi2022ccgg}, but may suffer from structural collapse or accumulate irreversible sampling errors in complex molecular graph spaces. 
% Recent advances integrate graph generative frameworks with diffusion models~\cite{bogetsimple,wang2024mixed}. Methods such as DiGress~\cite{vignac2022digress}, GDSS~\cite{jo2022score}, and EDM~\cite{hoogeboom2022equivariant} leverage forward–reverse diffusion processes to model graph topology. While all three adopt diffusion-based formulations, DiGress and GDSS operate primarily in graph space, whereas EDM focuses on equivariant diffusion in continuous 3D space. 
Recent advances integrate graph generative frameworks with diffusion and flow matching formulations~\cite{bogetsimple,wang2024mixed,qinmadeira2024defog}. Methods such as DiGress~\cite{vignac2022digress} and GDSS~\cite{jo2022score} leverage forward-reverse diffusion processes to model graph topology, while DeFoG~\cite{qinmadeira2024defog} formulates graph generation through discrete flow matching, decoupling the training and sampling stages to improve sampling flexibility and efficiency. 
However, these graph generators lack powerful guidance mechanisms, leading to large conditional variance and inconsistent intermediate representations. To overcome this limitation, our approach introduces a data-anchored diffusion process that leverages intrinsic mean reversion to provide adaptive restoration toward the target distribution.%, leading to more stable training and higher-quality molecular graph generation.
% To address this issue, our approach introduces a Vasicek-based, data-anchored diffusion process that ensures convergence to the target distribution, resulting in more stable training and higher-quality molecular generation.
% 分子天然地可以表示为图结构，其中节点代表原子、边代表化学键。图生成模型旨在通过学习节点与边的联合分布，直接在图空间中生成满足化学约束的新分子。这类方法关注于如何在生成过程中保持分子的价数、环结构、连通性等关键结构特征，从而保证化学合法性与结构可解释性。早期方法多采用自回归或变分图解码器生成节点和边，然而在复杂分子空间中，这类方法容易导致结构塌陷或局部最优。为克服这些问题，研究者提出了多种改进方案，如在图生成过程中引入稳定噪声机制、Ornstein–Uhlenbeck stochastic process动力学调制以及基于属性的条件引导，以提升采样稳定性与分子新颖性。近年来，图结构生成框架逐渐与扩散模型相结合，例如 DiGress、EDM\cite{hoogeboom2022equivariant}、GDSS 等模型通过在图空间定义扩散—逆扩散过程，实现了分子拓扑结构与三维几何约束的统一建模。这类模型不仅提高了生成分子的化学有效性，也为基于图结构的SMDC配体发现提供了新的可能。

% \subsection{Diffusion Bridge Models}
% Diffusion bridge models offer a principled framework for integrating prior knowledge by constructing stochastic transitions between endpoint distributions, proving effective for molecular optimization and image restoration.
Diffusion bridge models construct stochastic transitions between endpoint distributions and have been applied to molecular optimization and image restoration.
Diffusion Schr{\"o}dinger Bridge~\cite{de2021diffusion,kim2024discrete} formulates bridge learning as an entropy-regularized optimal transport problem solved via iterative proportional fitting, while Denoising Diffusion Bridge Models~\cite{zhou2023denoising} extend this idea by directly learning bridge scores to unify score-based and flow matching~\cite{20243116790693} methods. In image restoration, the Generalized Ornstein-Uhlenbeck Bridge (GOUB)~\cite{yue2024image} leverages the Doob’s $h$-transform~\cite{doob1984classical} to generalize SDE models. For graph generation, HOG-Diff~\cite{huang2026hog} incorporates higher-order topological structures such as cell complexes as a guiding mechanism into graph diffusion models, and relies on generalized OU diffusion bridges and Laplacian spectral diffusion to construct a coarse-to-fine hierarchical generation paradigm. GruM~\cite{jo2023graph} adopts an Ornstein-Uhlenbeck-bridge-based diffusion-mixing strategy to guide deterministic structural learning.  A key limitation of GruM~\cite{jo2023graph} is its temporally imbalanced generative dynamics. During early sampling, the endpoint-directed guidance can be weak relative to the stochastic influence inherited from the noise prior, delaying the movement of high-noise states toward the data manifold; near the target time, however, its bridge guidance coefficient becomes singular. 
VGD addresses this challenge by introducing an adaptive drift mechanism that smoothly regulates the restoring force throughout the trajectory, providing a more balanced transition from noise initialization to complex graph manifolds.

\section{VGD for Graph Generation}
% \section{Diffusion Bridge Process for Generative Modeling}
% We consider a sampling diffusion process using a Vasicek model, a special case of the Ornstein-Uhlenbeck process characterized by data-anchored dynamics and time-dependent variance control. In our adaptation for generative modeling, we reformulate the Vasicek model as a sampling diffusion process that connects a noise distribution to a target data distribution. Formally, the stochastic differential equation is defined by:
\subsection{Vasicek Graph Diffusion}
% We propose a direct-sampling model based on the Vasicek diffusion process~\cite{VASICEK1977177}, which enables straightforward generation of samples through forward diffusion from noise to data. 
We propose a direct-sampling model based on the Vasicek diffusion process~\cite{VASICEK1977177}, in which both training and sampling are formulated in a common noise-to-data direction, enabling a unified and naturally directed sampling process.
% 我们提出了一种基于 Vasicek 过程的直接采样扩散模型，该模型能够通过从噪声到数据的单次前向传播直接生成样本。
% Unlike conventional diffusion models that rely on reverse processes, 
Our VGD approach utilizes a data-anchored diffusion that directly transforms Gaussian noise at $t=0$ into target data samples at $t=T$. 
% 与依赖迭代逆过程的传统扩散模型不同，我们的方法利用均值回复扩散，直接将 $t=0$ 时的 Gaussian 噪声转换为 $t=T$ 时的目标数据样本。
Formally, the SDE is defined by:
\begin{align}
\mathrm{d}x_t 
&=f_t{\rm d}t+g_t{\rm d}w_t
% \notag\\&
= \gamma(x_T - x_t)\mathrm{d}t + g_t\mathrm{d}w_t
\label{eq:Vasicek}
\end{align}
where $x_t$ denotes the state variable at time $t \in [0,T]$, $x_T$ denotes the target data sample that serves as the reversion level, $\gamma > 0$ governs the strength of data reversion toward $x_T$, $f_t$ is the drift term, $g_t$ is a time-dependent diffusion term that controls the noise injection schedule, and $w_t$ is a standard Wiener process. 
% 其中 $x_t$ 表示时间 $t \in [0,T]$ 时的状态变量，$x_T$ 表示作为均值回归水平的目标数据样本，$\gamma > 0$ 控制均值回归到 $x_T$ 的强度，$g_t$ 是控制噪声注入计划的时间相关扩散系数，$w_t$ 是标准维纳过程。

The statistics of Eq.~\eqref{eq:Vasicek} are analytically tractable:
% Vasicek 方程的统计特性可以通过解析方法处理：
\begin{align}
\mu_t
% [x_t |x_0, x_T] 
&= e^{-\gamma t} x_0 + (1 - e^{-\gamma t}) x_T\,,\label{eq:mu} \\
\sigma_t^2
% [x_t |x_0, x_T] 
&= e^{-2\gamma t} \int_0^t e^{2\gamma s} g_s^2 \mathrm{d}s\,.\label{eq:sigma}
\end{align}
Our key insight emerges from analyzing the boundary behavior of Eq.~\eqref{eq:mu}: it represents an exponentially weighted interpolation between the initial noise $x_0$ and the target data $x_T$. At $t=0$, $\mu_0 = x_0 \sim \mathcal{N}(0,1)$, initializing the process with $x_0$. Crucially, under the condition of sufficiently large $\gamma$ and $T$, we observe that $\mu_T \approx x_T$, demonstrating that the process asymptotically converges to the target data distribution.
This convergence behavior, combined with the variance analysis from Eq.~\eqref{eq:sigma}, reveals a profound property. Specifically, when $\gamma$ and $T$ are sufficiently large, $\sigma_T^2 \approx 0$, and the target standard deviation $\sigma_T$ is much smaller than the discretization interval. This ensures that the target distribution collapses to a point mass centered at $x_T$, while the conditional mean follows data-anchored trajectory from the noise prior toward the target data. This property preserves stochastic exploration at intermediate times while ensuring stable graph discretization near the target endpoint.

Our principal contribution lies in establishing this diffusion behavior, which provides: (a) Exact Gaussian transitions throughout the entire trajectory, enabling efficient sampling and training; (b) Deterministic convergence to the target data distribution; and (c) Analytical tractability for both mean and variance, eliminating the need for numerical approximations as shown in Eqs.~\eqref{eq:mu} and \eqref{eq:sigma}.

% To the best of our knowledge, this represents the first formulation of a diffusion process that simultaneously maintains Gaussian transitional distributions while achieving asymptotically exact bridge behavior through principled parameter selection. 
This theoretical foundation provides a generative framework that combines stable training dynamics with efficient sampling. This property implies that the diffusion trajectory effectively ``bridges'' the standard Gaussian prior and the data manifold through a data-anchored path.
In other words, the process evolves from noise to data under a vanishing-variance constraint, ensuring a smooth and stable transition across all intermediate states.
% Details are provided in \textcolor{red}{Appendix}.

We initialize the sampling with \(x_0\sim\mathcal{N}(0, 1)\) and allow it to diffuse from \(t=0\) to \(t=T\), converging toward a target state \(x_T\) drawn from the data distribution. The diffusion process in Eq.~\eqref{eq:Vasicek} is constructed such that, starting from standard Gaussian noise, it gradually diffuses into the data distribution as $t \rightarrow T$. The VGD sampling algorithm is given by Algorithm~\ref{alg:VGDsampling}, where $\widehat{x}_T(\theta^\star)=\psi(x_t,t|\theta^\star)$ denotes a neural network $\psi$ of $x_T$ parameterized by the optimized $\theta^\star$.
% 我们初始化采样过程为 \(x_0 \sim \mathcal{N}(0, I) \)，并允许其从 \( t=0 \) 扩散到 \( t=T \)，最终收敛到从数据分布中抽取的终端状态 \( x_T \)。方程~\eqref{eq:vbd_bridge} 中的扩散过程构造如下：从标准高斯噪声开始，当 $t \rightarrow T$ 时，它逐渐过渡到数据分布。

\begin{algorithm}[!t]
\caption{VGD sampling.}\label{alg:VGDsampling}
\begin{algorithmic}[1]
\REQUIRE  {$\theta^\star$\,.}
\item[\textbf{Initialize:}] $x_0\sim\mathcal{N}(0, 1)$\,.
\FOR{$n \in[0,N-1]$}
\STATE $t=\frac{n}N,\Delta t=\frac1N$\,;
\STATE $z\sim \mathcal{N}(0, 1)$\,;
\STATE $\widehat{x}_T(\theta^\star)=\psi(x_t,t|\theta^\star)$\,;
\STATE $x_{t+\Delta t}=x_t+\gamma(\widehat{x}_T(\theta^\star) - x_t)\Delta t + g_t z\sqrt{\Delta t}$\,;
\ENDFOR
\ENSURE {A generated $x_T$\,.}
\end{algorithmic}
\end{algorithm}

\subsection{VGD Training}
During training, a neural network, $\widehat{x}_T(\theta)=\psi(x_t,t|\theta)$, learns to predict \( x_T \) from any intermediate noisy sample \( x_t \). 
These closed-form expressions enable consistent sampling from $p(x_t|x_0,x_T)=\mathcal N(x_t|\mu_t,\sigma_t^2)$, allowing us to generate training noisy samples \( x_t \).
% 这些闭式表达式能够从 $p(x_t|x_0,x_T)=\mathcal N(x_t|\mu_t,\sigma_t)$ 中高效且一致地采样，从而生成训练噪声样本 \( x_t \)。
% The neural network is trained to estimate \( x_T \) given \( x_t \).
Under the minimum mean square error (MMSE) criterion, this task corresponds to learning the conditional expectation \( \mathbb{E}[x_T|x_t] \).
% 神经网络经过训练，能够在给定 \( x_t \) 的情况下估计 \( x_T \)，这在最小均方误差准则下对应于学习条件期望 \( \mathbb{E}[x_T|x_t] \)。
% Since the target variance \(\sigma^2_{t=T}\) asymptotically approaches zero, we can replace the stochastic endpoint by its expectation and use the predicted mean to guide each sampling trajectory toward the learned data manifold.
Substituting this estimate into the data-anchored drift yields a learned marginal drift that guides the state toward the data distribution. As the conditional perturbation decreases near the target time, the endpoint estimate is expected to become increasingly reliable.

% 由于终端方差 \(\sigma_{t=T}\) 渐近趋于零，我们可以用其期望值替换随机端点，并在生成过程中使用预测均值进行确定性采样。

In our approach, we leverage VGD for generative modeling by learning the target $x_T$ through $\widehat{x}_T(\theta)=\psi(x_t,t|\theta)$. The training procedure operates as follows:
% 在我们的方法中，我们利用这种扩散桥进行生成建模，通过神经网络学习目标 $x_T$。训练过程如下：
\begin{enumerate}
    \item For each time $t$, we can generate noisy samples of $x_t$ using the known distribution $\mathcal N(x_t|\mu_t,\sigma_t^2)$.
    % 对于每个时间 $t$，我们可以使用已知的分布 $\mathcal N(x_t|\mu_t,\sigma_t)$ 生成 $x_t$ 的噪声样本。
    \item These samples $x_t$ are fed into $\widehat{x}_T(\theta)=\psi(x_t,t|\theta)$ that predicts the target $x_T$.
    % 这些样本 $x_t$ 被输入到一个神经网络中，该神经网络预测目标 $x_T$。
    \item From MMSE estimation theory, the optimal predictor for $x_T$ given $x_t$ is the conditional expectation $\mathbb{E}[x_T|x_t]$.
    % 根据最小均方误差估计理论，给定 $x_t$ 时 $x_T$ 的最优预测器是条件期望 $\mathbb{E}[x_T|x_t]$。
\end{enumerate}

The VGD training algorithm is given by Algorithm~\ref{alg:VGDtrain}. As the variance of VGD at the target time $T$ approaches zero, we can use $\mathbb{E}[x_T|x_t]$ as a surrogate for $x_T$ during sampling.

% 当过程在终止时间 $T$ 的方差接近于零时，我们可以利用这个条件期望作为采样期间实际 $x_T$ 的替代值。

%_________________
% \subsection{Noise Schedule and Conditional Variance}

% \begin{algorithm}[!t]
% \caption{VGD training.}\label{alg:VGDtrain}
% \begin{algorithmic}[1]
% \REQUIRE  {A training set $\mathcal D=\{x_T\}$\,;}
% \item[\textbf{Initialization:}] $\theta, \gamma, g_0$ and $g_T$\,;
% \FOR{$t \sim {\rm uniform}(0,1)$}
% \STATE $x_T\sim \mathcal{D}\,;$
% \STATE $x_0\sim\mathcal{N}(0,1)$\,;
% \STATE Calculate $\mu_t$ by Eq.~\eqref{eq:mu}\,;
% \STATE Calculate $\sigma_t^2$ by Eq.~\eqref{eq:conditional_variance}\,;
% \STATE $x_t\sim \mathcal{N}(\mu_t, \sigma^2_t)$\,;
% \STATE $\widehat{x}_T(\theta)=\psi(x_t,t|\theta)$\,;
% \STATE $\mathcal{L}_{\rm vgd}=\frac{\gamma^2}{g_t^2}\|x_T-\widehat{x}_T(\theta)\|^2$\,;
% \STATE Update $\theta$ by backpropagation of $\mathcal{L}_{\rm vgd}$\,;
% \ENDFOR
% \ENSURE {$\theta^\star=\theta$\,;}
% \end{algorithmic}
% \end{algorithm}
The design of the diffusion term \(g_t\) is crucial for controlling the stochasticity of the diffusion process. We employ a linear interpolation scheme for the squared diffusion  term:
\begin{equation}
g_t^2 = (1-t) g_0^2 + t g_T^2, \quad t \in [0,1]
\label{eq:g_schedule}
\end{equation}
where \(g_0\) and \(g_T\) are hyperparameters representing the prior and target diffusion strengths, respectively.

This linear parameterization offers several advantages: (a) Smooth transition ensures a gradual evolution from initial noise to target state; (b) Analytical tractability permits closed-form solutions for Eq.~\eqref{eq:sigma}; (c) Controllable stochasticity provides explicit control over noise injection throughout the diffusion trajectory; and (d) Intermediate stochastic exploration is retained while residual perturbations near the endpoint are suppressed.  Inspired by the classical parameterizations adopted in SDE models, namely the linear parameterization in VP-SDE and the exponential parameterization in VE-SDE, we compare these two choices within the proposed VGD framework in Appendix~\ref{app:var}. Results show that the linear parameterization consistently achieves better performance while maintaining analytical simplicity. Therefore, we adopt the linear diffusion term throughout this paper. 

With this parameterization, the conditional variance \(\sigma_t^2\) of Eq.~\eqref{eq:sigma} admits an analytical expression:
\begin{equation}
\sigma_t^2 = \frac{g_t^2}{2\gamma} + \frac{g_0^2 - g_T^2}{4\gamma^2} - e^{-2\gamma t} \left[ \frac{g_0^2}{2\gamma} + \frac{g_0^2 - g_T^2}{4\gamma^2} \right].
\label{eq:conditional_variance}
\end{equation}
During training, once \(\gamma\), \(g_0\) and \(g_T\) are fixed as hyperparameters, the conditional variance \(\sigma_t^2\) follows a deterministic trajectory dictated entirely by the time.% without numerical integration. %This analytical computability provides significant benefits: (a) Training stability eliminates variance estimation errors that could propagate through gradient computations; (b) Computational efficiency avoids expensive numerical integration during the training loop; and (c) Reproducibility ensures consistent variance profiles across different training runs.

The hyperparameters \(g_0\) and \(g_T\) are typically selected to balance exploration and convergence: \(g_0\) is set to provide sufficient initial stochasticity for covering the noise distribution, while \(g_T\) is chosen as a small positive value to ensure stable convergence to the target distribution while maintaining numerical stability across the entire trajectory.

% Eq.~\eqref{eq:conditional_variance} allows us to precisely control the diffusion process's stochastic properties while maintaining the theoretical guarantees of VGD

%_________________

The training objective is to minimize the KL divergence between the path measure of the true data-generating process \(p\) and the parameterized model \(q_\theta\). For two It\^o diffusion processes sharing the same diffusion  term \(g_t\), Girsanov's theorem~\cite{oksendal2013stochastic} provides an explicit form for this divergence:
\begin{align}
D_{\text{kl}}(p \| q_\theta) = \mathbb{E}_{p}\left[ \frac12 \int_0^T \left\| g_t^{-1} \left( f_t - f_\theta \right) \right\|^2  {\rm d}t \right] + C \notag
\end{align}
where \(C\) is a constant independent of \(\theta\), \(f_t\) is the drift of the true process, and \(f_\theta\) is the drift of the model.

For our proposed VGD, the SDE is defined as Eq.~\eqref{eq:Vasicek}. The true drift \(f_t\), which guides the process from \(x_0\) to the true target \(x_T\), is $f_t = \gamma(x_T - x_t)$. The model drift $f_\theta$, which uses the network's prediction is: $f_\theta = \gamma(\widehat{x}_T(\theta)- x_t)$.

Compute the difference between these drifts:
\begin{align}
f_t - f_\theta 
% = \gamma(x_T - x_t) - \gamma(\widehat{x}_T(\theta)- x_t) 
= \gamma(x_T - \widehat{x}_T(\theta))\,.\notag
\end{align}
Substituting this into the Girsanov formula yields:
\begin{align}
D_{\rm kl}(p \parallel q_\theta) 
% &= \mathbb{E}_{p}\left[ \frac{1}{2} \int_0^T \left\| g_t^{-1} \cdot \gamma (x_T - \widehat{x}_T(\theta)) \right\|^2  dt \right] + C \notag\\
&= \mathbb{E}_{p}\left[ \frac{1}{2} \int_0^T \frac{\gamma^2}{g_t^2} \| x_T - \widehat{x}_T(\theta)\|^2  {\rm d}t \right] + C\,.\notag
\end{align}

Therefore, minimizing the KL divergence is equivalent to minimizing the following loss function:
\begin{align}
\mathcal{L}_{\rm vgd} 
= \frac{\gamma^2}{g_t^2} \| x_T-\widehat{x}_T(\theta) \|^2\,. 
\label{eq:VGD_loss}
\end{align}

\begin{algorithm}[!t]
\caption{VGD training.}\label{alg:VGDtrain}
\begin{algorithmic}[1]
\REQUIRE  {A training set $\mathcal D=\{x_T\}$\,.}
\item[\textbf{Initialize:}] $\theta, \gamma, g_0$ and $g_T$\,.
\FOR{$t \sim {\rm uniform}(0,1)$}
\STATE $x_T\sim \mathcal{D}\,, x_0\sim\mathcal{N}(0,1)$;
\STATE Calculate $\mu_t$ by Eq.~\eqref{eq:mu}\,;
\STATE Calculate $\sigma_t^2$ by Eq.~\eqref{eq:conditional_variance}\,;
\STATE $x_t\sim \mathcal{N}(\mu_t, \sigma^2_t)$\,;
\STATE $\widehat{x}_T(\theta)=\psi(x_t,t|\theta)$\,;
\STATE $\mathcal{L}_{\rm vgd}=\frac{\gamma^2}{g_t^2}\|x_T-\widehat{x}_T(\theta)\|^2$\,;
\STATE Update $\theta$ by minimizing $\mathcal{L}_{\rm vgd}$\,;
\ENDFOR
\ENSURE {$\theta^\star=\theta$}\,.
\end{algorithmic}
\end{algorithm}
% Interpretation: 
% The coefficient \(\frac{\gamma^2}{g_t^2}\) is not arbitrary but is theoretically mandated. It acts as an optimal time-dependent weight, emphasizing prediction accuracy during low-noise phases (small \(g_t\)) and down-weighting it when the inherent stochasticity of the process is high. 
This result rigorously connects the MMSE criterion with the principle of minimizing distributional divergence via Girsanov's theorem. Although the training objective is derived from the principle of minimizing the path-wise KL divergence via Girsanov's theorem, the resulting loss function is a weighted mean-squared error. This ensures that the learned function \(\widehat{x}_T(\theta)=\psi(x_t, t|\theta)\) converges to the MMSE estimator for the endpoint, which is \(\mathbb{E}[x_T | x_t]\).

% VGD provides a computationally efficient approach for generative modeling: the model learns to approximate $\mathbb{E}[x_T|x_t]$, enabling the generation of high-quality samples by VGD from noise to data.
% Vasicek 扩散模型为生成建模提供了一种计算效率高的方法：该模型学习近似 $\mathbb{E}[x_T|x_t]$，从而能够通过 Vasicek 扩散从噪声生成高质量的样本。

\subsection{VGD for Graph Generation}
A graph with $N$ nodes is denoted as $G = (X, A)$, where $X \in \mathbb{R}^{N \times F}$ is the node feature matrix and $A \in \mathbb{R}^{N \times N}$ is the adjacency matrix~\cite{jiang2019semi}. 
% In molecular graphs, $X$ encodes atomic features, and $A$ represents the chemical bonds between atoms. Given a dataset of $M$ observed graphs $\mathcal{G} = {G_i}{i=1}^M$, 
The goal of graph generation is to learn the underlying distribution $p(G)$ and to sample new graphs $\widetilde{G}_T \sim p(G)$ that are statistically consistent with the training data.
% For molecular graphs, valid generation additionally requires respecting chemical valency rules, while also achieving high novelty and diversity.
% 我们将$N$个节点的图表示为$G=(X,A)$，$X\in\mathbb{R}^{N\times F}$代表节点特征矩阵，$A\in\mathbb{R}^{N\times N}$代表节点之间的加权邻接矩阵，在分子图任务中用于表示节点间化学键的存在与否以及化学建的类型。给定一组包含$M$个观测图的集合$G=\{G_i\}_{i=1}^M$,图生成任务的核心是学习这些观测图所服从的概率分布$p(G)$ ,并能从该学习到的分布中采样生成新的图$G_{new}$,且满足$G_{new}\sim p(G)$,即新生成的图在分布上与观测图保持一致。\cite{9490655}这些生成的图不仅应在统计分布上与观测图保持一致，还必须满足化学分子的价键规则，并具备新颖性和多样性，以探索更广阔的结构空间。

In our graph sampling approach, we adopt the VGD formulation to model the diffusion process as shown in Algorithm~\ref{alg:VGDsampling}. Starting from a random initialization, both $X_0$ and $A_0 \sim \mathcal{N}(0, 1)$, the graph diffusion over time is described by the following SDE:
\begin{equation}
\left\{
\begin{aligned}
\mathrm{d}X_t &= f_X(X_t,A_t,t)\,\mathrm{d}t + g_{X,t}\,\mathrm{d}W_{X,t}\\
\mathrm{d}A_t &= f_A(X_t,A_t,t)\,\mathrm{d}t + g_{A,t}\,\mathrm{d}W_{A,t}
\end{aligned}
\right.
\label{eq:VGD_graph_sde}
\end{equation}
where $f_X$ and $f_A$ are drift terms, 
$g_{X,t}$ and $g_{A,t}$ are diffusion  terms, 
and $W_{X,t}$ and $W_{A,t}$ denote independent standard Wiener processes. 
The drift functions are given by:
\begin{equation}
\begin{aligned}
f_X(X_t,A_t,t) &= \gamma_X \big(\widehat{X}_T(\theta^\star) - X_t\big) \\
f_A(X_t,A_t,t) &= \gamma_A \big(\widehat{A}_T(\theta^\star) - A_t\big)
\end{aligned}
\label{eq:VGD_graph_drift}
\end{equation}
% where $\widehat{x}_T(\theta^\star)$ and $\widehat{A}_T(\theta)^\star$ denote the model’s estimates of the data node features and adjacency matrix. 
where \(\widehat{X}_T(\theta^\star)\) and \(\widehat{A}_T(\theta^\star)\) denote the model’s predicted node features and adjacency matrix of the target graph from any intermediate noisy
sample $G_t$, respectively. Both predictors share the same parameter set $\theta^\star$, which is optimized as described in Algorithm~\ref{alg:VGDtrain} by minimizing the VGD loss defined in Eq.~\eqref{eq:VGD_loss}. The coefficients $\gamma_X$ and $\gamma_A$ control the drift strengths for the node and edge components.

We adopt the VGD to model the graph diffusion process. VGD defines a continuous-time stochastic process over both node features and adjacency matrices, where the drift terms guide the noisy graph representations toward the target distribution, while the diffusion terms control the injected stochasticity. Compared with purely discrete generative models, this continuous formulation allows for more stable training.
After sampling, we follow GDSS~\cite{jo2022score} and GruM~\cite{jo2023graph} to discretize the generated node and edge representations.

\subsection{Insights into VGD}
The effectiveness of diffusion-based graph generation is fundamentally determined by how efficiently useful structural information emerges along the diffusion trajectory. Figure~\ref{fig:var_GDSS} reveals two important observations. Existing graph diffusion models either experience relatively high uncertainty during diffusion or exhibit delayed growth of SNR, making informative graph structures difficult to recover in early stages. In contrast, VGD maintains a smoothly evolving conditional variance while increasing the SNR earlier, allowing meaningful structural information to emerge throughout the trajectory rather than only near the target stage.
\begin{figure}[!t]
\centering
% \hspace*{-8mm}  
\includegraphics[width=0.96\columnwidth]{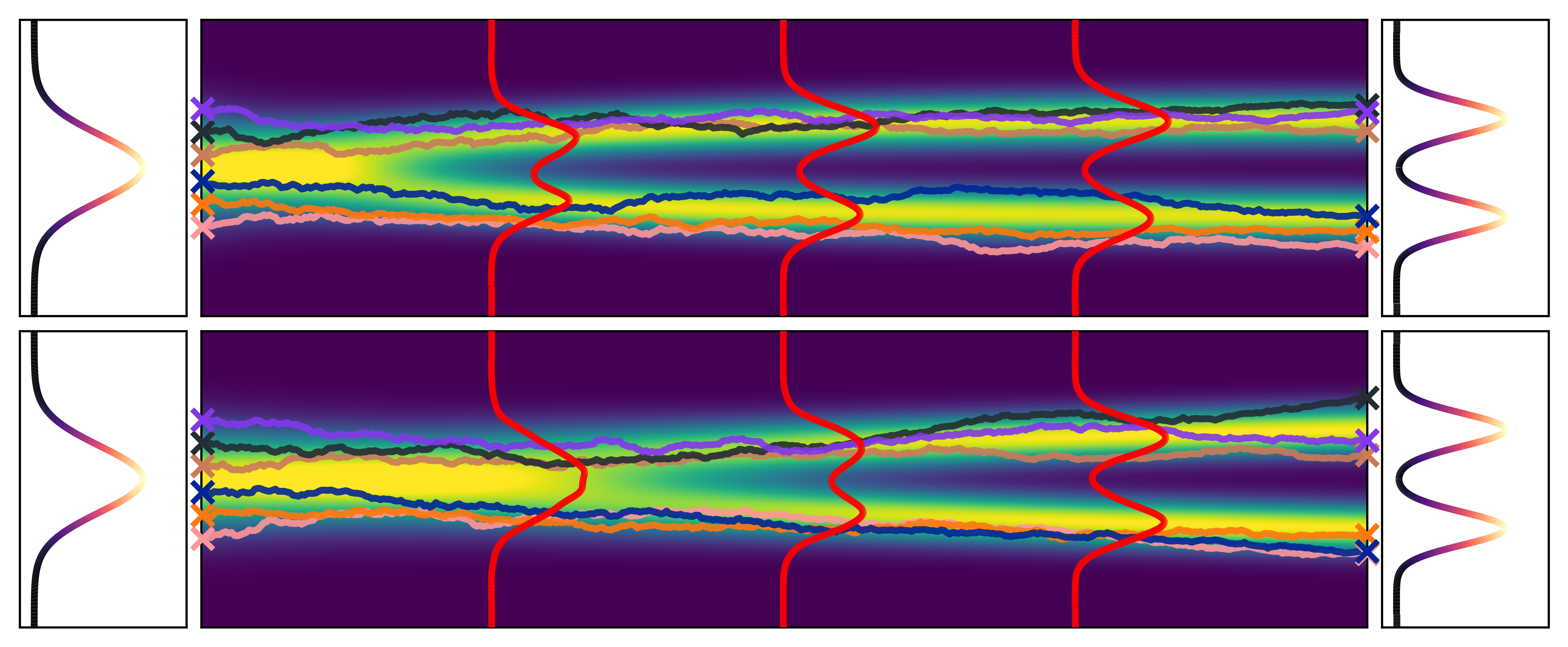}
\caption{Evolution of probability densities from a unimodal Gaussian prior to a bimodal target distribution.
Compared with the conventional OU bridge (bottom), the proposed VGD (top) separates the two target modes substantially earlier, demonstrating a more effective transition from noise to the underlying data manifold.
% Sampling trajectories of VGD (top) and the OU-bridge baseline (bottom). VGD separates the two target modes earlier under the evaluated settings.
}\label{fig:compare_path}
\end{figure}

To further understand this behavior, Figure~\ref{fig:compare_path} visualizes the evolution of a one-dimensional bimodal distribution. The conventional OU bridge remains concentrated around the intermediate region for a large portion of the trajectory before separating into the two target modes near the end. By contrast, VGD exhibits substantially earlier mode separation, indicating that the sampling trajectories commit to their target manifolds much sooner. This earlier discrimination enables subsequent sampling steps to focus on refining local graph structures rather than continuously resolving large-scale uncertainty.

These observations can be explained by the dynamics of VGD. The data-anchored drift continuously steers the stochastic process toward the target graph, while the controlled diffusion process prevents excessive uncertainty from overwhelming the informative signal. Their combination yields a diffusion trajectory that is both analytically tractable and computationally efficient.%, providing a principled explanation for the empirical improvements observed across graph generation benchmarks.

\section{Experiment}
\subsection{Experimental Setting}
% Our neural network $\psi$ references the Transformer-based architecture of GruM~\cite{jo2023graph}. Generation is performed using Euler-Maruyama with $1000$ function evaluations. The node and edge diffusion processes use the linear squared-diffusion schedule defined in Eq.~\eqref{eq:g_schedule}, with dataset-specific values of $\gamma,g_0$, and $g_T$. 
% The SDE hyperparameters $(\gamma,g_0,g_T)$ are set to $(9,0.4,0.2)$ for QM9, $(13,0.8,0.1)$ for ZINC250k, $(6,0.6,0.2)$ for Planar, and $(11,0.8,0.2)$ for SBM. For simplicity, we use the same hyperparameters for the node-feature and adjacency processes. This parameter sharing is an implementation choice rather than a requirement of our formulation, and component-specific settings can be used when necessary. Algorithm~\ref{alg:VGDsampling} presents the base sampler and complete dataset-specific setting are provided in Appendix.

Our neural network $\psi$ references the Transformer-based architecture of GruM~\cite{jo2023graph}. Node, edge, and global attributes are projected into a latent space and jointly updated through stacked Transformer layers with multi-head attention, which models node-edge interactions and uses aggregated global information to refine local representations. Finally, residual MLPs map the latent features back to their original dimensions.

 For simplicity, we use the same hyperparameters for the node-feature and adjacency processes. This parameter sharing is an implementation choice rather than a requirement of our formulation, and component-specific settings can be used when necessary. The node and edge diffusion processes use the linear squared-diffusion schedule defined in Eq.~\eqref{eq:g_schedule}, with dataset-specific values of $\gamma,g_0$, and $g_T$. The SDE hyperparameters of VGD, $\gamma, g_0$, and $g_T$, are set to $9,0.4,0.2$ for QM9, $13,0.8,0.1$ for ZINC250k, $6,0.6,0.2$ for Planar, and $11,0.8,0.2$ for SBM, respectively. Complete implementation and training details are provided in Appendix~\ref{app:implementation_details}.

The discretization follows GruM. For generic graphs, edge prediction is thresholded at 0.5. For molecular graphs, node prediction is similarly thresholded at 0.5. The continuous edge output is in the range of [0, 3] and mapped to discrete bond orders based on specific intervals: $[0,0.5) \rightarrow 0$ (no bond), $[0.5, 1.5) \rightarrow 1$, $[1.5,2.5) \rightarrow 2$, and $[2.5,3] \rightarrow 3$.

% \begin{table}[!t]
% \centering
% \resizebox{0.6\linewidth}{!}{
% {
% \setlength{\tabcolsep}{4pt} 
% \begin{tabular}{lcccccc}
% \toprule
% {Dataset} & $\gamma$ & $g_0$ & $g_1$\\
% \midrule

% QM9 & 9 & 0.4 & 0.2  \\
% ZINC250k & 13 & 0.8 & 0.1 \\
% Planar & 6 & 0.6 & 0.2 \\
% SBM & 11 & 0.8 & 0.2 \\
% % \rowcolor[HTML]{D9EAF7}
% \bottomrule
% \end{tabular}
% }
% }
% \caption{SDE hyperparameters used for VGD.}
% \label{tab:hyperparameters}
% \end{table}

\subsection{Molecule Generation}
\subsubsection{Datasets.}
% We conduct a comprehensive evaluation on two widely used molecular graph datasets: QM9~\cite{ramakrishnan2014quantum} and ZINC250k~\cite{irwin2012zinc}. The QM9 dataset consists of 133,885 small organic molecules, each containing up to nine atoms from a set of five atom types. It serves as a standard benchmark for molecular generation tasks, commonly used to assess the performance of models in terms of validity, uniqueness, and diversity of the generated molecular structures. The ZINC250k dataset, a subset of the ZINC compound library, contains 249,455 molecules that satisfy drug-likeness constraints, with each molecule comprising up to 38 atoms from ten atom types. Molecules in both datasets feature three types of chemical bonds: single, double, and triple. Each molecule is represented as a 2D graph structure, where nodes correspond to atoms and edges denote chemical bond types. Compared to QM9, ZINC250k encompasses a broader range of structural diversity and higher chemical complexity.

We conduct a comprehensive evaluation on two widely adopted molecular graph benchmarks: QM9~\cite{ramakrishnan2014quantum} and ZINC250k~\cite{irwin2012zinc}. The QM9 dataset comprises 133,885 small organic molecules with up to nine heavy atoms, specifically carbon, nitrogen, oxygen, and fluorine. The ZINC250k dataset, a drug-like subset of the ZINC library, exhibits higher chemical complexity and structural diversity, containing 249,455 molecules with up to 38 heavy atoms across ten distinct atom types. Following standard protocols in molecular generation, we adopt established practices for both graph representation and dataset splitting. In this structure, nodes correspond to heavy atoms encoded by their specific atom types, while edges represent chemical connections categorized as single, double or triple bonds. Hydrogen atoms are explicitly removed during preprocessing. 

\subsubsection{Metrics.}
To evaluate generation performance, we employ four standard metrics: (1) \texttt{Valid} measures the ratio of chemically validity molecules generated without post-correction, verified via RDKit valency checks. (2) \texttt{FCD}~\cite{preuer2018frechet} quantifies the distributional distance in the chemical feature space, where lower values indicate closer property distributions to the real data. (3) \texttt{NSPDK MMD}~\cite{costa2010fast} utilizes a subgraph-based kernel to assess the structural distribution discrepancy between molecular graphs; lower values signify higher structural consistency. (4) \texttt{Scaf}. measures the similarity of scaffold structures between generated and real molecules, with higher scores representing closer alignment with reference scaffolds.
% 为了全面评估模型的分子生成性能，我们采用了多种主流指标，包括 Validity、FCD\cite{preuer2018frechet}、NSPDK MMD\cite{costa2010fast} 和 Scaf. (Scaf.)。
% Validity 衡量生成分子在没有做任何矫正前是否为化学意义上有效的结构，该指标反映了模型在生成阶段对化学约束的学习能力。FCD用于衡量生成分子与真实分子在特征空间中的分布差异。较低的 FCD 值表示生成分子在化学性质分布上更接近真实数据。NSPDK MMD衡量生成分子图与真实分子图在结构分布上的相似性。NSPDK 是一种基于子图的图核方法，通过捕捉局部邻域结构的统计特征来评估图分布差异。较小的 MMD 值表示生成分子在图结构层面与真实数据更加一致。Scaf. (Scaf.) 用于衡量生成分子与真实分子在骨架结构层面的相似度。分子骨架由去除支链和取代基后的核心环状结构构成，该指标能够反映模型在生成新颖分子时对基本化学骨架的保持或创新能力。较高的Scaf.表明生成分子在结构核心上与真实分子更为接近。

\subsubsection{Baselines.}
We compare our method against representative state-of-the-art baselines across various generative paradigms. These include MoFlow~\cite{zang2020moflow}, a normalizing flow-based model, and GraphAF~\cite{shi2020graphaf}, which employs an autoregressive structure for efficient molecular generation. GraphDF~\cite{luo2021graphdf} extends this line of work by introducing discrete latent variables to enhance generative flexibility. 
% Among score-based models, we compare against EDP-GNN~\cite{niu2020permutation}, which features permutation invariance, and GDSS~\cite{jo2022score}, a model that excels at graph distribution modeling. Unlike earlier diffusion models, DiGress~\cite{vignac2022digress} generates graph structures with categorical node and edge attributes through a discrete denoising diffusion mechanism, and GruM~\cite{jo2023graph} leverages a diffusion-mixing strategy to further enhance generation quality.
Among score-based models, we compare against EDP-GNN~\cite{niu2020permutation}, which features permutation invariance, and GDSS~\cite{jo2022score}, a model that excels at graph distribution modeling. Among more recent diffusion-based models, DiGress~\cite{vignac2022digress} generates graph structures with categorical node and edge attributes through a discrete denoising diffusion mechanism, while GruM~\cite{jo2023graph} leverages a diffusion-mixing strategy to further enhance generation quality. DeFoG~\cite{qinmadeira2024defog} formulates graph generation through discrete flow matching to decouple the training and sampling stages. HOG-Diff~\cite{huang2026hog} further incorporates higher-order structures into OU diffusion bridges to construct a coarse-to-fine hierarchical generation.
% 为了验证我们方法的有效性，我们在多个基准数据集上与以下分子图生成模型进行对比:MoFlow \cite{zang2020moflow} 是一种基于归一化流的分子图生成模型，GraphAF \cite{shi2020graphaf} 采用自回归结构实现高效的分子生成，GraphDF \cite{luo2021graphdf} 进一步引入离散潜变量以增强生成灵活性。基于分数生成的模型包括 EDP-GNN \cite{niu2020permutation}，其具备置换不变性特性，以及 GDSS \cite{jo2022score}，在图分布建模上表现优异。扩散模型方面，DiGress \cite{vignac2022digress} 通过离散去噪扩散机制生成带有分类节点和边属性的图结构，而 GruM \cite{jo2023graph} 则利用扩散混合策略进一步提升生成质量。

\subsubsection{Results.}
As shown in Table~\ref{tab:qm9_results}, our approach VGD demonstrates exceptionally strong and comprehensive performance on QM9. Specifically, VGD shows a clear advantage in the more challenging metrics of \texttt{FCD}, \texttt{NSPDK}, and \texttt{Scaf}., which better assess the quality and diversity of the generated molecules. Relative to the strongest baseline for each metric, VGD reduces \texttt{FCD} by 10.5\% and \texttt{NSPDK} by 33.3\%, and improves \texttt{Scaf}. by 0.0062. This suggests that our method does more than maximize validity; it achieves a superior balance, generating molecules that are both chemically valid and rich in structural diversity and relevance. Notably, while the GruM~\cite{jo2023graph} model achieves the highest validity score, standing at 99.86\%, our approach VGD attains a highly competitive result of 99.35\%.

% As illustrated in Table~\ref{tab:zinc250k_results}, our method achieves SOTA results on all four evaluation metrics in ZINC250k. Most notably, VGD reduces the \texttt{FCD} score from the previous best of 1.633 to 1.210 and improves \texttt{Scaf}. from 0.5278 to 0.5560, highlighting its strong capability in modeling and recovering complex molecular property distributions. Additionally, the high \texttt{Valid} and \texttt{Scaf}. scores further confirm the model's powerful generative capabilities.

As shown in Table~\ref{tab:zinc250k_results}, VGD achieves the best results across all four metrics on ZINC250k. In particular, its \texttt{FCD} of 1.210 and \texttt{NSPDK} of 0.0005 indicate that the generated molecules closely match the reference distribution in both chemical feature space and graph-structural space. Meanwhile, the validity of 99.83\% shows that this distributional fidelity is achieved without compromising chemical feasibility, while the \texttt{Scaf}. score of 0.5560 demonstrates strong alignment with the scaffold distribution of real molecules. The simultaneous improvements across these complementary metrics are particularly notable on ZINC250k, where larger molecular graphs and more diverse atom types introduce substantially more complex node-edge dependencies than in QM9.

\begin{table}[!t]
\centering
\resizebox{\linewidth}{!}{
{
\setlength{\tabcolsep}{4pt} 
\begin{tabular}{lcrcc}
\toprule
{Method} & \texttt{Valid} (\%)$\uparrow$ & \texttt{FCD}$\downarrow$ & \texttt{NSPDK}$\downarrow$ & \texttt{Scaf}.$\uparrow$ \\
\midrule
% MoFlow~\cite{zang2020moflow} & 91.36 & 4.467 & 0.0169 & 0.1447 \\
% GraphAF~\cite{shi2020graphaf} & 74.43 & 5.625 & 0.0267 & 0.3046 \\
% GraphDF~\cite{luo2021graphdf} & 93.88 & 10.928 & 0.0046 & 0.3778 \\
% EDP-GNN~\cite{niu2020permutation} & 47.52 & 2.680 & 0.0046 & 0.2302 \\
% GDSS~\cite{jo2022score} & 95.72 & 2.900 & 0.0023 & 0.6983 \\
% DiGress~\cite{vignac2022digress} & 99.59 & 0.300 & 0.0006 & 0.9353 \\
% GruM~\cite{jo2023graph} & {99.86} & 0.124 & 0.0004 & 0.9357 \\
MoFlow & 91.36 & 4.467 & 0.0169 & 0.1447 \\
GraphAF & 74.43 & 5.625 & 0.0207 & 0.3046 \\
GraphDF & 93.88 & 10.928 & 0.0636 & 0.0978 \\
EDP-GNN & 47.52 & 2.680 & 0.0046 & 0.3270 \\
GDSS & 95.72 & 2.900 & 0.0023 & 0.6983 \\
DiGress & 99.59 & 0.300 & 0.0006 & 0.9353 \\
GruM & 99.86 & 0.124 & 0.0004 & 0.9357 \\
DeFoG & 97.12 & 0.114 & 0.0004 & 0.9408 \\
HOG-Diff & 98.74 & 0.172 & 0.0003 & - \\
\midrule
% \rowcolor[HTML]{D9EAF7}
% \rowcolor{gray!10}
\textbf{VGD} (Ours) & 99.35 & {0.102} & {0.0002} & {0.9470} \\
\bottomrule
\end{tabular}
}
}
\caption{Comparison of VGD with baselines on QM9. 
% The best results are in \textbf{bold}.
}
\label{tab:qm9_results}
\end{table}

%我们的实验结果表明，在训练过程中截断数据端的扩散过程可以显著提升模型性能。我们将这种性能提升归因于模型终端方差的有效正则化。在VGD，当扩散过程接近终端时间 T 时，状态 x(t) 会收敛到真实数据流形的附近，但其条件方差会变得非常小。这种极小的方差严重限制了模型的探索能力，从而抑制了生成样本的多样性。通过实施数据端截断，我们有效地放宽了 VGD 模型施加的严格终端约束。这神经网络学习一个更鲁棒的生成模型，使其能够探索更广泛的数据空间。我们的实验证实，这种策略不仅增强了生成样本的新颖性，而且显著提高了其有效性。

\subsection{General-Purpose Graph Generation}
\subsubsection{Datasets.}
We evaluate our approach VGD on two generic graph datasets, Planar and Stochastic Block Model (SBM). The Planar dataset is composed of a series of randomly generated planar graphs, where each graph is topologically guaranteed to be free of crossing edges, which satisfies the planarity constraint. Due to its regular topological structure and well-defined statistical properties, Planar is often regarded as a standard testbed for validating the structural fidelity and stability of graph generation models. The SBM dataset consists of synthetic graphs produced by the Stochastic Block Model (SBM). The model generates graphs with distinct community structure by using different edge probabilities for edges within versus between communities. This dataset is employed to assess the model's ability to capture high-level structural patterns. In contrast to Planar, SBM inherently focuses more on macroscopic topological regularities and global generative consistency.
% 我们在两个通用数据集上对模型进行了评估，包括Planar和Stochastic Block Model (SBM).Planar数据集由一系列随机生成的平面图组成，每个图在拓扑结构上保证无交叉边，从而满足平面可嵌入性约束。由于其拓扑结构规则且统计特性明确，Planar常被视为验证图生成模型结构保真性与稳定性的标准测试场景。SBM数据集是一类基于随机块模型的合成图数据集，通过在不同社区之间设定不同的连边概率来生成具有显著社群结构的图。该数据集可用于评估模型在捕获高层次结构模式方面的表现。与 Planar相比，SBM 更关注图的宏观拓扑规律与全局生成一致性.

\begin{table}[!t]
\centering
\resizebox{\linewidth}{!}{
{
\setlength{\tabcolsep}{4pt} 
\begin{tabular}{lcrcc}
\toprule
{Method} & \texttt{Valid} (\%)$\uparrow$ & \texttt{FCD}$\downarrow$ & \texttt{NSPDK}$\downarrow$ & \texttt{Scaf}.$\uparrow$ \\
\midrule
% MoFlow~\cite{zang2020moflow} & 63.11 & 20.931 & 0.0455 & 0.0133 \\
% GraphAF~\cite{shi2020graphaf} & 68.47 & 16.023 & 0.0442 & 0.0672 \\
% GraphDF~\cite{luo2021graphdf} & 90.61 & 33.546 & 0.1770 & 0.0000 \\
% EDP-GNN~\cite{niu2020permutation} & 82.97 & 16.737 & 0.0485 & 0.0000 \\
% GDSS~\cite{jo2022score} & 97.01 & 14.656 & 0.0195 & 0.0467 \\
% DiGress~\cite{vignac2022digress} & 94.99 & 3.482 & 0.0021 & 0.4163 \\
% GruM~\cite{jo2023graph} & 98.34 & 2.207 & 0.0016 & 0.5278 \\
MoFlow & 63.11 & 20.931 & 0.0455 & 0.0133 \\
GraphAF  & 68.47 & 16.023 & 0.0442 & 0.0672 \\
GraphDF  & 90.61 & 33.546 & 0.1770 & 0.0000 \\
EDP-GNN  & 82.97 & 16.737 & 0.0485 & 0.0000 \\
GDSS  & 97.01 & 14.656 & 0.0195 & 0.0467 \\
DiGress  & 94.99 & 3.482 & 0.0021 & 0.4163 \\
GruM & 98.34 & 2.207 & 0.0016 & 0.5278 \\
DeFoG & 99.22 & 2.123 & 0.0022 & 0.4245 \\ 
HOG-Diff & 98.56 & 1.633 & 0.0010 & - \\
\midrule
% \rowcolor[HTML]{D9EAF7}
% \rowcolor{gray!10}
\textbf{VGD} (Ours) & {99.83} & {1.210} & {0.0005} & {0.5560} \\
% \textbf{Ours(truncation  + var. )} & \textbf{99.47} & \textbf{1.139} & \textbf{0.0007} & 
% \textbf{0.5040} \\
% \textbf{Ours(truncation)} & \textbf{99.45} & \textbf{1.145} & \textbf{0.0008} & \textbf{0.5171} \\
\bottomrule
\end{tabular}
}
}
\caption{Comparison of VGD with baselines on ZINC250k. 
% The best results are in \textbf{bold}.
}
\label{tab:zinc250k_results}
\end{table}

\subsubsection{Metrics.}
To comprehensively evaluate the model's performance on graph generation tasks, we adopted five mainstream metrics for structural consistency and generation quality, including Degree (\texttt{Deg}.), Clustering (\texttt{Clus}.), 4-node orbit counts (\texttt{Orbit}), Spectral (\texttt{Spec}.), and Validity-Uniqueness-Novelty (\texttt{V}.\texttt{U}.\texttt{N}.). \texttt{Deg}. measures the discrepancy in the node degree distribution between the generated and reference graphs, reflecting the model's ability to capture local connectivity patterns. \texttt{Clus}. quantifies differences in the distribution of clustering coefficients, capturing connectivity within node neighborhoods and local community structure. \texttt{Orbit} compares distributions of local subgraphs, evaluating the model’s capacity to learn higher-order structural motifs. The \texttt{Spec}. metric, based on the graph Laplacian spectrum, is used to measure the discrepancy in global structural features, reflecting the graph's overall connectivity and structural complexity. Finally, \texttt{V}.\texttt{U}.\texttt{N}. provides a composite evaluation of the validity, uniqueness, and novelty of the generated samples. For the Planar dataset, a graph is considered valid only if it is connected and satisfies the planarity constraint, as verified using the Boyer--Myrvold algorithm.
% 为全面评估模型在图生成任务中的性能，我们采用了五项主流的结构一致性与生成质量指标，包括Degree（\texttt{Deg}.）、Clustering（\texttt{Clus}.）、Orbit、Spectral（\texttt{Spec}.） 和 Validity–Uniqueness–Novelty（\texttt{V}.\texttt{U}.\texttt{N}.）。\texttt{Deg}. 衡量生成图与真实图在节点度分布上的差异，反映模型在捕获局部连接模式方面的能力。\texttt{Clus}.衡量生成图与真实图在聚类系数分布上的差异。该指标反映节点邻域之间的连通性和局部群聚结构。Orbit 指标用于评估生成图与真实图在子结构或轨道分布上的差异。它通过比较不同类型局部子图在生成样本与真实样本中的统计分布，衡量模型在学习高阶结构模式方面的能力。\texttt{Spec}.指标基于图拉普拉斯谱，用于衡量生成图与真实图在整体结构特征上的差异。该指标反映图的全局连通性和结构复杂度。\texttt{V}.\texttt{U}.\texttt{N}.综合评估生成样本的有效性、唯一性与新颖性。

\subsubsection{Baselines.}
To validate the generalization capability of the proposed method on general-purpose graph generation tasks, we conducted a comparison with several representative baseline models on the Planar and SBM datasets. Among autoregressive models, GraphRNN~\cite{you2018graphrnn} relies on recurrent neural networks, whereas GRAN~\cite{liao2019efficient} utilizes attention-based Graph Neural Networks (GNNs).
SPECTRE~\cite{pmlr-v162-martinkus22a} models the graph generation process in the spectral domain. ConGress~\cite{vignac2022digress} is a continuous diffusion model. The other diffusion models, including EDP-GNN~\cite{niu2020permutation}, GDSS~\cite{jo2022score}, and GruM~\cite{jo2023graph}, have been introduced in the preceding sections.

\begin{table}[!t]
\centering
\resizebox{\linewidth}{!}{
{
\setlength{\tabcolsep}{3pt} 
\begin{tabular}{lccccr}
\toprule
{Method} & \texttt{Deg}.$\downarrow$ & \texttt{Clus}.$\downarrow$ & \texttt{Orbit}$\downarrow$ & \texttt{Spec}.$\downarrow$ & \texttt{V}.\texttt{U}.\texttt{N}.$\uparrow$ \\
\midrule
GraphRNN & 0.0049 & 0.2779 & 1.2543 & 0.0459 & 0.0~~~ \\
GRAN & 0.0007 & 0.0426 & {0.0009} & 0.0075 & 0.0~~~ \\
SPECTRE & 0.0005 & 0.0785 & 0.0012 & 0.0112 & 25.0~~~ \\
EDP-GNN & 0.0044 & 0.3187 & 1.4986 & 0.0813 & 0.0~~~ \\
GDSS & 0.0041 & 0.2676 & 0.1720 & 0.0370 & 0.0~~~ \\
ConGress & 0.0048 & 0.2728 & 1.2950 & 0.0418 & 0.0~~~ \\
% DiGress~\cite{vignac2022digress} & 0.0003 & 0.0372 & 0.0009 & 0.0106 & 75.0~~~ \\
% DiGress~\cite{vignac2022digress} & \textbf{0.0003} & 0.0372 & \textbf{0.0009} & \textbf{0.0106} & 75.0 \\
% GruM & 0.0005 & 0.0353 & 0.0009 & 0.0062 & 90.0 \\
% GruM_{ckpt} & 0.0006 & \textbf{0.0342} & 0.0001 & 0.0066 & \textbf{90.0} \\
GruM & 0.0007 & 0.0489 & 0.0048 & 0.0075 & 82.5~~~ \\
% GruM(none) & 0.0008 & 0.0347 & 0.0000 & 0.0083 & 92.50 \\
% DeFoG & 0.0005 & 0.0501 & 0.0006 & 0.0072 & 99.5~~~ \\
\midrule
% \textbf{Ours(four)} & 0.0004 & \textbf{0.0305} & \textbf{0.0000} & 0.0062 & 90.00 \\
% \textbf{Ours(none)} & 0.0004 & 0.0341 & \textbf{0.0000} & 0.0099 & \textbf{97.50} \\
% \rowcolor[HTML]{D9EAF7}
% \rowcolor{gray!10}
\textbf{VGD} (Ours) & {0.0002} & {0.0368} & 0.0035 & {0.0052} & {90.0}~~~ \\
\bottomrule
\end{tabular}
}
}
\caption{Comparison of VGD with baselines on Planar. 
% he comparison of our method with baselines on 
% The best results are in \textbf{bold}.
}
\label{tab:planar_results}
\end{table}

% As shown in Table~\ref{tab:planar_results}, our proposed VGD model exhibits superior overall performance on Planar . In particular, it achieves the lowest errors on the Degree, Clustering, and Spectral metrics, indicating that the generated graphs align closely with the ground-truth distribution in both structural and spectral aspects. Although the  score is marginally higher than that of the best baseline, the overall generation quality remains competitive. Moreover, VGD achieves a \texttt{V}.\texttt{U}.\texttt{N}. score of 90.0, outperforming advanced diffusion-based models such as GruM~\cite{jo2023graph} in terms of generation validity and diversity. These results clearly validate the capability of our approach VGD to capture both local and global topological dependencies, while maintaining high diversity and validity in the generated graphs.
\subsubsection{Results.}
As presented in Table~\ref{tab:planar_results}, our proposed VGD model demonstrates state-of-the-art performance on the challenging task of planar graph generation. Specifically, VGD achieves the best results across nearly all metrics that evaluate structural similarity. It obtains the best result in \texttt{Deg}., \texttt{Clus}., and \texttt{Spec}., indicating a superior capability in capturing both local and global graph properties compared to all baselines. While the \texttt{Orbit} score in GRAN~\cite{liao2019efficient} is competitive, the most significant advantage of our method is highlighted by the \texttt{V}.\texttt{U}.\texttt{N}. metric. VGD achieves a \texttt{V}.\texttt{U}.\texttt{N}. score of 90.0, substantially outperforming the previous best model. This demonstrates that VGD not only generates graphs with high structural fidelity but also excels in producing a diverse set of valid, unique, and novel planar graphs.

% 如表~\ref{tab:planar_results} 所示，我们的方法在Planar数据集上的各项指标均表现出优异性能。模型在Degree和Spectral两项指标上取得了最低误差，表明生成图在结构和谱特征方面与真实分布高度一致。尽管Clustering指标略高于最优基线模型，但整体生成质量依然保持在极高水平。此外，我们的模型在 \texttt{V}.\texttt{U}.\texttt{N}.指标上达到90.00，与GruM~\cite{jo2023graph} 等先进扩散模型持平。这一结果充分证明了所提出方法在捕获图的局部与全局拓扑特征方面的有效性，同时兼具良好的多样性与生成有效性。

% As illustrated in Table~\ref{tab:sbm_results}, our VGD model achieves the best performance on four out of five metrics, including Degree, Clustering, \texttt{Orbit}, and Spectral, demonstrating strong capability in modeling both local and global graph structures. Although its \texttt{V}.\texttt{U}.\texttt{N}. score is slightly lower than that of GruM~\cite{jo2023graph}, VGD still ranks second in this metric, indicating competitive generation validity and diversity. Overall, VGD achieves a well-balanced performance across all evaluation metrics on SBM .

The results on SBM, shown in Table~\ref{tab:sbm_results}, further validate the robustness and effectiveness of our VGD model for general-purpose graph generation. VGD consistently outperforms or matches the best-performing baselines across all structural metrics. It achieves the best scores for \texttt{Clus}., \texttt{Orbit}, and \texttt{Spec}., and ties with GruM~\cite{jo2023graph} for the best \texttt{Deg}. score. This consistent, top-tier performance across a comprehensive set of statistics underscores VGD's strength in accurately modeling the underlying community structures characteristic of SBM graphs. In terms of generative diversity, VGD obtains a high \texttt{V}.\texttt{U}.\texttt{N}. score of 72.5, which is highly competitive and surpasses most existing methods, trailing only slightly behind GruM. In aggregate, these results confirm that VGD is a powerful and versatile graph generation model, capable of producing high-fidelity graphs for datasets with distinct structural properties.

\begin{table}[!t]
\centering
\resizebox{\linewidth}{!}{
{
\setlength{\tabcolsep}{3pt} 
\begin{tabular}{lccccr}
\toprule
{Method} & \texttt{Deg}.$\downarrow$ & \texttt{Clus}.$\downarrow$ & \texttt{Orbit}$\downarrow$ & \texttt{Spec}.$\downarrow$ & \texttt{V}.\texttt{U}.\texttt{N}.$\uparrow$ \\
\midrule
% GraphRNN~\cite{you2018graphrnn} & 0.0055 & 0.0584 & 0.0785 & 0.0065 & 5.0~~~ \\
% GRAN~\cite{liao2019efficient} & 0.0113 & 0.0553 & 0.0540 & 0.0054 & 25.0~~~ \\
% SPECTRE~\cite{martinkus2022spectre} & 0.0015 & 0.0521 & 0.0412 & 0.0056 & 52.5~~~ \\
% EDP-GNN~\cite{niu2020permutation} & 0.0011 & 0.0552 & 0.0520 & 0.0070 & 35.0~~~ \\
% GDSS~\cite{jo2022score} & 0.0212 & 0.0646 & 0.0894 & 0.0128 & 5.0~~~ \\
% ConGress~\cite{vignac2022digress} & 0.0273 & 0.1029 & 0.1148 & - & 0.0~~~ \\
% % DiGress~\cite{vignac2022digress} & 0.0013 & 0.0498 & 0.0434 & 0.0400 & 74.0~~~ \\
% GruM~\cite{jo2023graph} & {0.0006} & 0.0498 & 0.0455 & 0.0051 & {80.0}~~~ \\
GraphRNN & 0.0055 & 0.0584 & 0.0785 & 0.0065 & 5.0~~~ \\
GRAN & 0.0113 & 0.0553 & 0.0540 & 0.0054 & 25.0~~~ \\
SPECTRE & 0.0015 & 0.0521 & 0.0412 & 0.0056 & 52.5~~~ \\
EDP-GNN & 0.0011 & 0.0552 & 0.0520 & 0.0070 & 35.0~~~ \\
GDSS & 0.0212 & 0.0646 & 0.0894 & 0.0128 & 5.0~~~ \\
ConGress & 0.0273 & 0.1029 & 0.1148 & - & 0.0~~~ \\
GruM & {0.0006} & 0.0498 & 0.0455 & 0.0051 & {80.0}~~~ \\
% DeFoG & 0.0006 & 0.0517 & 0.0566 & 0.0054 & {90.0}~~~ \\
\midrule
% \rowcolor[HTML]{D9EAF7}
% \rowcolor{gray!10}
% \textbf{VGD} (Ours) & \textbf{0.0002} & \textbf{0.0498} & \textbf{0.0351} & \textbf{0.0044} & 67.5~~~ \\
\textbf{VGD} (Ours) & {0.0006} & {0.0494} & {0.0361} & {0.0044} & {72.5}~~~ \\
\bottomrule
\end{tabular}
}
}
\caption{Comparison of VGD with baselines on SBM. 
% The best results are in \textbf{bold}.
}
\label{tab:sbm_results}
\end{table}

\subsection{Ablation on the Number of Sampling Steps}
We investigate the influence of the number of sampling steps on VGD using the ZINC250k dataset. As shown in Table~\ref{tab:sampling_steps}, the generation quality improves steadily as the sampling trajectory becomes finer. Increasing the number of sampling steps from 25 to 200 consistently improves all evaluation metrics, reducing the FCD from 13.670 to 1.489 while increasing molecular validity from 95.97\% to 99.90\%. Using 1000 sampling steps yields the best overall performance reported.% in the paper.

\begin{table}[!t]
\centering
\resizebox{\linewidth}{!}{
\begin{tabular}{rrrrr}
% \begin{tabularx}{0.48\textwidth}<br/>{l *{8}{>{\centering\arraybackslash}X}}
\toprule
% \textbf{Dataset} &
Timesteps &
\texttt{Valid} (\%)~$\uparrow$ &
\texttt{FCD}~$\downarrow$ &
\texttt{NSPDK}~$\downarrow$ &
\texttt{Scaf}.~$\uparrow$ \\
\midrule

% ZINC250k&
 % 25~~~~~~   & 13.670 & 0.0185 & 95.97~~~~~~  & 0.2246 \\
 % 50~~~~~~   & 4.438  & 0.0062 & 99.50~~~~~~  & 0.3986 \\
 % 100~~~~~~  & 2.101  & 0.0021 & 99.79~~~~~~ & 0.4870 \\
 % 200~~~~~~  & 1.489  & 0.0011 & 99.90~~~~~~  & 0.5407 \\
 % 1000~~~~~~ & 1.210  & 0.0005 & 99.83~~~~~~ & 0.5560 \\

25~~~~~~   & 95.97~~~~~~ & 13.670 & 0.0185~~ & 0.2246~~ \\
50~~~~~~   & 99.50~~~~~~ &  4.438 & 0.0062~~ & 0.3986~~ \\
100~~~~~~  & 99.79~~~~~~ &  2.101 & 0.0021~~ & 0.4870~~ \\
200~~~~~~  & 99.90~~~~~~ & 1.489 & 0.0011~~ & 0.5407~~ \\
1000~~~~~~ & 99.83~~~~~~ & 1.210 & 0.0005~~ & 0.5560~~ \\

\bottomrule
\end{tabular}
}
% \end{tabularx}
\caption{VGD different sampling steps on ZINC250k.}
\label{tab:sampling_steps}
\end{table}

More importantly, VGD reaches competitive performance substantially earlier than existing diffusion bridges. With only 200 sampling steps, VGD already surpasses the 1000-step performance of GruM~\cite{jo2023graph} across all four evaluation metrics, demonstrating that the proposed diffusion dynamics require significantly fewer sampling transitions to approach the target graph distribution.

This observation is consistent with the analysis in Figures~\ref{fig:var_GDSS} and \ref{fig:compare_path}. Since VGD exhibits a faster increase in the SNR during diffusion, informative structural signals become recoverable earlier along the sampling trajectory. Consequently, fewer sampling steps are required before the generated samples become close to the target distribution, leaving more computation for subsequent structural refinement.

\section{Conclusion}
In this paper, we propose VGD, a novel diffusion model for graph generation. Leveraging a data-anchored property, VGD enables a smooth transformation from Gaussian noise to structured graphs. 
% This process is analytically tractable and admits a closed-form solution. 
% Notably, VGD maintains a coordinated temporal relationship between drift and variance, preserving sufficient stochastic exploration for structural diversity while allowing the data-directed drift to guide the trajectory more reliably as uncertainty decreases. 
% Unlike conventional diffusion models that mainly rely on stochastic perturbation, 
VGD explicitly coordinates the evolution of mean, variance, and SNR. The designed dynamics maintain controlled variance changes for stable intermediate states while rapidly increasing SNR at early diffusion stages, allowing the model to enter an informative regime earlier and allocate more sampling steps toward fine-grained structural refinement. Beyond its empirical effectiveness, VGD reveals interesting connections between different generative paradigms. The evolution of its mean follows an exponentially weighted linear interpolation between the initial noise and target data, sharing a similar perspective with flow matching in constructing smooth transport trajectories. Meanwhile, the variance gradually vanishes toward the data endpoint, resembling the convergence behavior of diffusion bridges and enabling effective alignment with the target data distribution. Interestingly, these properties naturally emerge from an OU process with a data-anchored drift. %, suggesting that carefully designed stochastic dynamics can bridge the advantages of flow-based transport, diffusion bridges, and stochastic diffusion models.
Extensive experiments on molecular and general-purpose graph benchmarks show that VGD achieves competitive validity and structural fidelity over leading diffusion models such as DiGress and GruM. These results highlight that explicitly controlling mean-variance dynamics is crucial for high-fidelity and stable graph diffusion.

{
    \small
    \bibliographystyle{ieeenat_fullname}
    \bibliography{main}
}
\clearpage
\clearpage
\appendix
\setcounter{secnumdepth}{2}
% \onecolumn
\begin{center}
    {\LARGE \bfseries Appendix} 
\end{center}
% \twocolumn
\section{Derivation of $\mu_t$ and $\sigma_t^2$ in Eqs.~(2) and (3)
% ~\eqref{eq:mu} and \eqref{eq:sigma}
}\label{sec:VGD}
\noindent
We derive the closed-form solution of the VGD defined by the linear SDE:
\begin{equation}
{\rm d}x_t = \gamma (x_T - x_t)\, {\rm d}t + g_t\, {\rm d}w_t\tag{1}
\end{equation}
which can be equivalently written as
\begin{equation}
{\rm d}x_t + \gamma x_t\, {\rm d}t = \gamma x_T\, {\rm d}t + g_t\, {\rm d}w_t .
\end{equation}

\paragraph{Step 1: Multiplying both sides by $e^{\gamma t}$ yiel{\rm d}s}
\begin{equation}
e^{\gamma t} {\rm d}x_t + \gamma e^{\gamma t} x_t\, {\rm d}t
= \gamma x_T e^{\gamma t}\, {\rm d}t + e^{\gamma t} g_t\, {\rm d}w_t.
\end{equation}
By applying the product rule for semimartingales, the left-hand side is recognized as the total differential:
\begin{equation}
{\rm d}( e^{\gamma t} x_t)
= \gamma x_T e^{\gamma t}\, {\rm d}t + e^{\gamma t} g_t\, {\rm d}w_t.
\end{equation}

\paragraph{Step 2: Integrating over time from $0$ to $t$}
\begin{equation}
\int_0^t {\rm d}(e^{\gamma s} x_s)
= \int_0^t \gamma x_T e^{\gamma s}\, {\rm d}s
+ \int_0^t e^{\gamma s} g_s\, {\rm d}w_s .
\end{equation}
Applying the fundamental theorem of calculus:
\begin{equation}
\int_0^t {\rm d}(e^{\gamma s} x_s)
= e^{\gamma t} x_t - e^{\gamma \cdot 0} x_0
= e^{\gamma t} x_t - x_0 .
\end{equation}
Then,
\begin{equation}
e^{\gamma t} x_t - x_0 = \gamma x_T \int_0^t e^{\gamma s}\, {\rm d}s
+ \int_0^t e^{\gamma s} g_s\, {\rm d}w_s.
\end{equation}

\paragraph{Step 3: Solving for $x_t$}
The deterministic integral is
\begin{equation}
\int_0^t e^{\gamma s}\, {\rm d}s = \frac{e^{\gamma t} - 1}{\gamma}.
\end{equation}
Substituting into the equation above gives:
\begin{equation}
e^{\gamma t} x_t - x_0
= x_T \left(e^{\gamma t} - 1\right)
+ \int_0^t e^{\gamma s} g_s\, {\rm d}w_s .
\end{equation}
Multiplying both sides by $e^{-\gamma t}$ yiel{\rm d}s the explicit solution:
\begin{equation}
x_t = e^{-\gamma t} x_0
+ \left(1 - e^{-\gamma t}\right) x_T
+ e^{-\gamma t} \int_0^t e^{\gamma s} g_s\, {\rm d}w_s,
\label{eq:xt}
\end{equation}
which completes the derivation. Consequently, the mean and variance of the VGD Process are given by:
\begin{align}
\mu_t
&= e^{-\gamma t} x_0 + (1 - e^{-\gamma t}) x_T\,,\tag{2}\\
\sigma_t^2
&= e^{-2\gamma t} \int_0^t e^{2\gamma s} g_s^2 \mathrm{d}s\,.\tag{3}
\end{align}

\section{Analytical Score Function of VGD}
Although VGD is trained by predicting the target graph rather than the score function, it naturally admits an analytical score estimator. Then, if predictor-corrector (PC) sampling~\cite{song2021scorebased} or Langevin correction is desired, the score function can be computed directly from the predicted data without training an additional score network.

Given the conditional Gaussian distribution
\begin{equation}
p(x_t|x_T)=\mathcal{N}(x_t\mid \beta_t x_T,v_t),
\end{equation}
where
\(
\beta_t=1-e^{-\gamma t},
\)
and
\(
v_t=e^{-2\gamma t}+\sigma_t^2,
\)
the conditional score is
\begin{align}
\nabla_{x_t}\log p(x_t|x_T)
=
v_t^{-1}
(\beta_t x_T-x_t).
\label{eq:cond_score}
\end{align}

Using Fisher's identity,
\begin{align}
\nabla_{x_t}\log p_t(x_t)
&=
\int
p(x_T|x_t)
\nabla_{x_t}\log p(x_t|x_T)
\,{\rm d}x_T \notag\\
&=
v_t^{-1}
\left(
\beta_t
\mathbb{E}[x_T|x_t]
-
x_t
\right).
\end{align}

Since the target state predictor of VGD is trained to approximate the posterior expectation,
\(
\widehat{x}_T(x_t,t;\theta^\star)
\approx
\mathbb{E}[x_T|x_t],
\)
the marginal score can be estimated as
\begin{equation}
% \boxed{
\nabla_{x_t}\log p_t(x_t)
\approx
v_t^{-1}
\left(
\beta_t
\widehat{x}_T(x_t,t;\theta^\star)
-
x_t
\right).
% }
\label{eq:vdm_score}
\end{equation}

Based on Eq.~(\ref{eq:vdm_score}), a Langevin corrector can be directly incorporated into the reverse diffusion process. At each diffusion step, the state is refined by
\begin{equation}
x_t^{k+1}
=
x_t^{k}
+
\epsilon
v_t^{-1}
\left(
\beta_t
\widehat{x}_T(x_t^{k},t;\theta^\star)
-
x_t^{k}
\right)
+
\sqrt{2\epsilon}\,z^{k},
\end{equation}
where $k$ denotes the corrector iteration, $\epsilon$ is the step size, and $z^{k}\sim\mathcal N(0,I)$. This correction step iteratively moves the sample toward the true marginal distribution while preserving stochasticity, making the proposed VGD readily compatible with existing Predictor-Corrector sampling frameworks.

\section{VGD Close-Form Variance}
\label{app:var}
Inspired by the two classical parameterizations used in score-based diffusion models, namely the linear diffusion coefficient in VP-SDEs and the exponential diffusion coefficient in VE-SDEs, we investigate both parameterizations within the proposed VGD framework.
\subsection{Variance under Linear Diffusion Coefficient}
Starting from Eq.~\eqref{eq:sigma} and using the linear diffusion coefficient,
\begin{equation}\label{eq:g_linear}
g_t^2 = (1-t)g_0^2 + t g_T^2 = g_0^2 + (g_T^2-g_0^2)t,
\end{equation}
set \(a:=g_0^2\) and \(b:=g_T^2-g_0^2\). Then
\[
\begin{aligned}
\int_0^t e^{2\gamma s} g_s^2 \,\mathrm{d}s
&= a\int_0^t e^{2\gamma s}\,\mathrm{d}s
+ b\int_0^t s e^{2\gamma s}\,\mathrm{d}s \\
&= a\frac{e^{2\gamma t}-1}{2\gamma}
+ b\left(\left[ s \frac{e^{2\gamma s}}{2\gamma} \right]_0^t - \int_0^t \frac{e^{2\gamma s}}{2\gamma} \,\mathrm{d}s\right) \\
&= a\frac{e^{2\gamma t}-1}{2\gamma}
+ b\left(\frac{t e^{2\gamma t}}{2\gamma} - \frac{1}{2\gamma} \left[ \frac{e^{2\gamma s}}{2\gamma} \right]_0^t\right) \\
&= a\frac{e^{2\gamma t}-1}{2\gamma}
+ b\left(\frac{t e^{2\gamma t}}{2\gamma} - \frac{e^{2\gamma t}-1}{4\gamma^2}\right) \\
\end{aligned}
\]
where we used integration by parts for the second integral. Multiplying by \(e^{-2\gamma t}\) gives
\[
\begin{aligned}
\sigma_t^2
&= a\frac{1-e^{-2\gamma t}}{2\gamma}
+ b\left(\frac{t}{2\gamma} - \frac{1-e^{-2\gamma t}}{4\gamma^2}\right) \\
&= \frac{a+b t}{2\gamma}
\; -\; \frac{b}{4\gamma^2}
\; -\; e^{-2\gamma t}\!\left(\frac{a}{2\gamma} - \frac{b}{4\gamma^2}\right) \\
&= \frac{g_t^2}{2\gamma}
+ \frac{g_0^2 - g_T^2}{4\gamma^2}
- e^{-2\gamma t}\!\left(\frac{g_0^2}{2\gamma} + \frac{g_0^2-g_T^2}{4\gamma^2}\right).
\end{aligned}
\]

\subsection{Variance under Exponential Diffusion Coefficient}
We also consider $g_t=g_0(\frac{g_1}{g_0})^t$, set $k=\ln(\frac{g_1}{g_0})$, we have:
\[
\begin{aligned}
\sigma_t^2
&= e^{-2\gamma t}\int_0^t e^{2\gamma s} g_0^2 e^{2ks}\,\mathrm{d}s\\
&= g_0^2 e^{-2\gamma t} \int_0^t e^{2(\gamma + k)s}\mathrm{d}s\\
&= \frac{g_0^2}{2(\gamma + k)} \left( e^{2kt} - e^{-2\gamma t} \right)\\
&= \frac{g_0^2}{2\left(\gamma + \ln\frac{g_1}{g_0}\right)} \left[ \left(\frac{g_1}{g_0}\right)^{2t} - e^{-2\gamma t} \right].
\end{aligned}
\]

To investigate whether the performance of VGD depends on the choice of diffusion-coefficient parameterization, we compare the proposed linear parameterization with the exponential counterpart while keeping all remaining components unchanged, including the data-anchored coefficient \(\gamma\), diffusion coefficients \((g_0,g_T)\), network architecture, training strategy, and sampling procedure.

Table~\ref{tab:schedule_ablation} reports the experimental results. On QM9, both parameterizations achieve comparable performance, with the proposed linear parameterization producing slightly better FCD and scaffold scores while maintaining identical NSPDK performance. On the more challenging ZINC250k benchmark, the linear parameterization consistently outperforms the exponential counterpart across all evaluation metrics, yielding notably lower FCD and higher scaffold similarity.
\begin{table}[!t]
\centering
\resizebox{\linewidth}{!}{
\begin{tabular}{lcccc}
\toprule
\textbf{Schedule} & %\textbf{Method} &
\texttt{FCD}~$\downarrow$ &
\texttt{NSPDK}~$\downarrow$ &
\texttt{Valid} (\%)~$\uparrow$ &
\texttt{Scaf.}~$\uparrow$ \\
\midrule
QM9 & &&&\\
~~~~~Linear & 0.102 & 0.0002 & 99.35 & 0.9470 \\
~~~~~Exponential & 0.103 & 0.0002 & 99.60 & 0.9413 \\

\midrule

ZINC250k & &&&\\
~~~~~Linear & 1.210 & 0.0005 & 99.83 & 0.5560 \\
~~~~~Exponential & 1.692 & 0.0010 & 99.78 & 0.4949 \\
\bottomrule
\end{tabular}
}
\caption{Comparison of two diffusion-term schedules.}
\label{tab:schedule_ablation}
\end{table}

These results suggest that the proposed linear diffusion coefficient is sufficient to achieve competitive generative performance while preserving analytical simplicity. More importantly, it enables a smooth evolution of the marginal variance without introducing additional nonlinear hyperparameters, making it well suited for analyzing the effect of the proposed Vasicek dynamics independently of schedule-specific tuning. 
Although more flexible parameterizations can also be adopted, they introduce additional hyperparameters and make it difficult to isolate the contribution of the proposed data-anchored dynamics. Then, we adopt the linear diffusion coefficient throughout the paper as a simple yet effective design that preserves analytical tractability while achieving strong empirical performance.

\section{Comparison with OU Diffusion Bridges}
The proposed VGD shares the same objective as existing OU diffusion bridges instantiated by GruM, namely constructing a stochastic trajectory that connects a Gaussian prior to the target graph distribution. The fundamental difference lies in the diffusion dynamics. Conventional OU bridges constructed through Doob's \(h\)-transform employ a time-dependent drift,
\[
\eta_{\rm GruM}(x_t,t)
=
\alpha\sigma_t^2x_t
+
\frac{\sigma_t^2}{v_t}
\left(
\frac{\widehat{x}_T}{u_t}-x_t
\right),
\]
where the guidance strength is jointly determined by \(u_t\) and \(v_t\). In contrast, VGD adopts a simple data-anchored drift,
\[
\eta_{\rm VGD}(x_t,t)
=
\gamma(\widehat{x}_T-x_t),
\]
whose magnitude varies naturally with the distance between the current state and the predicted target.

For the VP-based OU bridge, the guidance coefficient exhibits highly unbalanced temporal behavior. During early diffusion, \(u_t\) is close to zero while \(v_t\) is large, resulting in a weak drift toward the predicted data. Consequently, the early trajectory is dominated by stochastic diffusion and receives little guidance from the target distribution. 
% Near the target stage, however, \(u_t\rightarrow1\) and \(v_t\rightarrow0\), causing the drift term to increase rapidly and leading to stiff reverse dynamics with large numerical sensitivity.
Near the terminal time, $u_t\rightarrow1$, while $v_t$ decreases approximately linearly with the remaining time $T-t$. Consequently, the data-guidance coefficient $\sigma_t^2/(u_tv_t)$ grows approximately as $1/(T-t)$, increasingly amplifying small data-prediction and discretization errors and thereby producing stiff generative dynamics with high numerical sensitivity.

By contrast, the proposed VGD maintains an effective guidance mechanism throughout the entire diffusion process. Since the drift is directly proportional to the distance \(|\widehat{x}_T-x_t|\), stronger guidance is naturally provided during early diffusion when the latent state is still far from the predicted target, while the guidance gradually weakens as the trajectory approaches the data manifold. This adaptive behavior eliminates the singular drift of OU bridges and produces a considerably smoother reverse trajectory.

The different drift dynamics also explain the empirical observations in Figures~\ref{fig:var_GDSS} and \ref{fig:compare_path}. Continuous data guidance enables VGD to preserve a lower uncertainty while increasing the signal-to-noise ratio from much earlier diffusion stages. As a result, probability mass separates toward different target modes earlier, allowing subsequent sampling steps to focus on refining graph structures instead of overcoming large stochastic uncertainty.

The preceding analysis suggests that VGD and GruM distribute data-directed guidance differently over the sampling trajectory. To examine whether this temporal difference is also reflected in molecular generation, we track FCD, NSPDK, and scaffold similarity at uniformly spaced checkpoints along a fixed 1,000-step sampling trajectory on ZINC250k. This setting differs from the sampling-step ablation in the main text, where the total number of discretization steps is varied while every trajectory is integrated over the complete interval from noise to data. Here, the horizontal axis instead denotes the current step index within the same 1,000-step sampling trajectory. 

\begin{figure*}[!t]
\centering
% \hspace*{-8mm}  
\includegraphics[width=2\columnwidth]{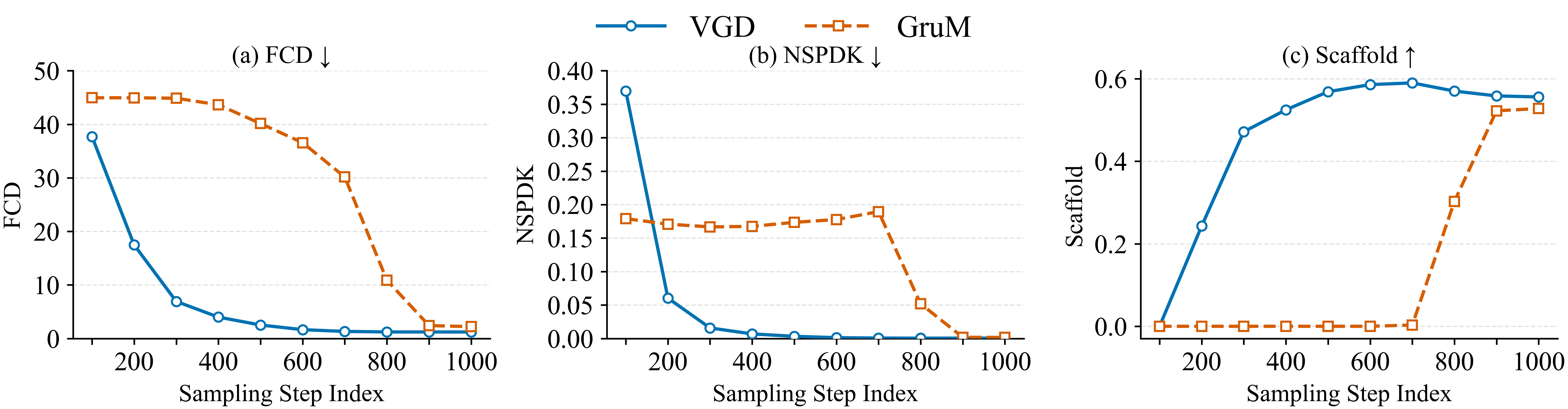}
\caption{Evolution of molecular-generation metrics along a fixed 1,000-step sampling trajectory on ZINC250k. Compared with GruM, VGD achieves favorable FCD, NSPDK, and scaffold similarity much earlier in the sampling process, whereas GruM exhibits substantial improvements only in the late stages as the trajectory approaches the data distribution.
% Evolution of molecular-generation metrics along a fixed 1,000-step sampling trajectory on ZINC250k. Compared with GruM, VGD reaches favorable FCD, NSPDK, and scaffold similarity values at earlier sampling stages, while GruM exhibits pronounced changes in these metrics mainly near endpoint, that is, the target data.
}\label{fig:compare_metrics}
\end{figure*}

% As shown in Figure~\ref{fig:compare_metrics}, VGD improves substantially earlier across all three distribution-sensitive metrics. Its FCD and NSPDK decrease rapidly during the first half of the trajectory, while the scaffold score begins to increase at an early stage and approaches a high-quality regime after approximately 500-600 steps. In contrast, the NSPDK and scaffold scores of GruM remain nearly unchanged throughout most of the trajectory, and its FCD decreases only gradually before all three metrics improve abruptly during the final sampling stages. This delayed transition is consistent with the weak early data guidance and rapidly increasing terminal guidance analyzed above.  
As shown in Figure~\ref{fig:compare_metrics}, VGD improves much earlier across all three distribution-sensitive metrics than GruM. Both FCD and NSPDK decrease rapidly during the early sampling stages, while the scaffold similarity begins to increase soon after generation starts and reaches a stable high-quality regime after approximately 500--600 sampling steps. In contrast, GruM exhibits only limited progress over most of the trajectory: FCD decreases slowly, whereas NSPDK and scaffold similarity remain nearly unchanged until the final sampling stages, where all three metrics improve rapidly. This behavior is consistent with the diffusion dynamics analyzed in the main text, where weak early guidance delays the recovery of meaningful graph structures.
% The curves further show that VGD enters an informative regime well before reaching the endpoint. After approximately 600--700 steps, its FCD and NSPDK are already close to their final values, such that the remaining trajectory mainly performs fine-grained structural refinement. The slight fluctuation in scaffold similarity after this stage also indicates that additional sampling steps do not monotonically improve every individual metric; rather, the primary distributional and structural recovery has already occurred. In contrast, GruM depends more heavily on late-stage updates to attain a comparable level of distributional alignment. These trajectory-level results therefore provide direct empirical evidence that the data-anchored drift enables useful molecular structures to emerge earlier instead of concentrating most of the recovery near the target data.
More importantly, the trajectories reveal that VGD approaches the target data distribution substantially earlier. After roughly 600-700 sampling steps, both FCD and NSPDK have already converged close to their final values, indicating that the remaining sampling steps are primarily devoted to refining local structural details rather than recovering the overall data distribution. In contrast, GruM relies heavily on the final stages of sampling to achieve comparable distributional alignment. These results provide direct empirical evidence that the data-anchored drift enables VGD to recover informative molecular structures much earlier, leaving a larger fraction of the sampling trajectory available for fine-grained structural refinement.

In essence, OU bridges achieve endpoint matching through a time-varying guidance field, whereas VGD achieves the same objective through distance-aware data-anchored dynamics. This redesign of the reverse dynamics fundamentally changes the diffusion trajectory, leading to earlier data commitment, smoother uncertainty evolution, and more stable graph generation.

\section{Implementation Details}
\label{app:implementation_details}
Our neural network $\psi$ references the Transformer-based architecture of GruM~\cite{jo2023graph}, which can effectively model the complex interactions between node, edge, and global features. First, the raw node, edge, and global attributes are projected into a latent space. The hidden layers consist of stacked Transformer layers with multi-head self-attention, enabling the model to capture both local structural details and global graph dependencies across multiple scales. In each hidden layer, a multi-head attention mechanism facilitates the interaction between node and edge features. Global features are dynamically updated by aggregating the statistical information of node and edge features, and are used to refine local representations. Finally, the latent features are mapped back to their original dimensions through a multi-layer perceptron with residual connections.

\begin{table*}[!t]
\centering
\resizebox{0.7\linewidth}{!}{
\begin{tabular}{lcccc}
\toprule
& {Planar} & {SBM} & {QM9} & {ZINC250k} \\
\midrule
Node hidden dimension        & 256 & 256 & 256 & 256 \\
Edge hidden dimension        & 64  & 64  & 128 & 64 \\
Transformer layers           & 8   & 8   & 8   & 9 \\
Attention heads              & 8   & 8   & 8   & 8 \\
Node feed-forward dimension  & 256 & 256 & 256 & 256 \\
Edge feed-forward dimension  & 64  & 64  & 256 & 128 \\
Time feed-forward dimension  & 256 & 256 & 128 & 128 \\
Time embedding dimension     & 64  & 64  & 64  & 64 \\
Output MLP layers            & 2   & 2   & 2   & 2 \\
Activation function          & ReLU & ReLU & ReLU & ReLU \\
Normalization                & LayerNorm& LayerNorm & LayerNorm & LayerNorm \\
\bottomrule
\end{tabular}
}
\caption{Network Architecture.}
\label{tab:general_model_hyper}
\end{table*}

\begin{table}[t]
\centering
\resizebox{\linewidth}{!}{
\begin{tabular}{lcccc}
\toprule
Parameter & {Planar} & {SBM} & {QM9} & {ZINC250k} \\
\midrule
Learning rate      & 4.0e-4 & 2.0e-4 & 2.0e-4 & 2.0e-4 \\
Weight decay       & 1.0e-12 & 1.0e-12 & 1.0e-12 & 1.0e-12 \\
Batch size         & 128 & 32 & 2048 & 512 \\
Training epochs    & 30000 & 30000 & 1000 & 500 \\
Gradient clipping  & 1.0 & 1.0 & 1.0 & 1.0 \\
EMA decay          & 0.999 & 0.999  & 0.999 & 0.999 \\
Corrector steps    & 1  & 0 & 0 & 0 \\
Generated samples  & 40 & 40 & 10,000 & 10,000 \\
\bottomrule
\end{tabular}
}
\caption{Training and sampling setting.}
\label{tab:experiment_config}
\end{table}
\subsection{Network Architecture and Setting}
In our experiments, we observe that the hyperparameters $\gamma$, $g_0$, and $g_T$ directly impact model performance, with $\gamma$ being the most critical factor. Specifically, a small $\gamma$ weakens both the data-directed drift and variance damping, leaving the trajectory noise-dominated and delaying SNR growth. Conversely, an excessively large $\gamma$ induces overly rapid contraction and stiff dynamics, suppressing useful stochastic exploration and increasing sensitivity to early data-prediction and discretization errors. Therefore, a moderate $\gamma$ is required to balance exploration, convergence speed, and sampling stability. The parameters $g_0$ and $g_T$ primarily regulate the stochasticity at the beginning and end of the trajectory, respectively. A larger $g_0$ promotes early exploration, whereas a small but positive $g_T$ suppresses residual perturbations near the data endpoint while maintaining numerical stability. Therefore, these parameters jointly balance early exploration, directed convergence, and terminal generation accuracy. 

These findings indicate that although the Planar and SBM datasets contain more nodes than QM9 and ZINC250k, they benefit from a smaller $\gamma$, which corresponds to a slower convergence rate. We attribute this to the higher complexity of constraints in molecular datasets compared to generic graphs. Molecular graphs require strict adherence to valency rules; thus, a larger coefficient is necessary to prevent the model from learning invalid or redundant information. Additionally, we find that even within a specific domain, datasets with larger graph sizes generally require a larger optimal $\gamma$. The architecture and dataset-specific experimental configurations are summarized in Tables~\ref{tab:general_model_hyper} and~\ref{tab:experiment_config}, respectively.

Moreover, the model performance typically follows a unimodal curve: it improves with increasing $\gamma$ until an optimal point, after which further increases lead to performance degradation. This is likely because an excessively large $\gamma$ prevents the model from learning the underlying structure effectively from the noisy states, resulting in misaligned guidance during the sampling process. We observe similar characteristics for $g_0$ and $g_T$. Consequently, the search landscape for these key parameters is well-behaved, making the optimal configuration relatively easy to locate.

\subsection{General-Purpose Graph Generation}
Following SPECTRE~\cite{pmlr-v162-martinkus22a} and GruM~\cite{jo2023graph}, we evaluate our model on synthetic datasets to assess the ability to capture topological structures:
\begin{itemize} [leftmargin=*]
\item \textbf{Planar Graphs:} The dataset consists of 200 synthetic graphs, each containing 64 nodes. A generated graph is considered valid if it is both connected and planar.
\item \textbf{SBM Graphs}: The dataset comprises 200 synthetic Stochastic Block Model graphs. The number of communities is uniformly sampled from [2,5], and the number of nodes per community is sampled from [20,40]. The probabilities for inter-community and intra-community edges are set to 0.05 and 0.3, respectively.
\end{itemize} 
We strictly adhere to the standard evaluation protocol established in SPECTRE~\cite{pmlr-v162-martinkus22a} and DiGress~\cite{vignac2022digress}, utilizing identical data splits and evaluation procedures. We report baseline results either directly from the original literature or by reproducing them using official open-source implementations with default hyperparameters.

\subsection{Molecule Generation}
We evaluate the quality of generated molecules on two standard benchmarks, QM9 and ZINC250k, following the experimental settings established in GruM~\cite{jo2023graph}. Both datasets are modeled with three edge types representing single, double, and triple bonds. To ensure a strictly fair comparison, we directly utilize the pre-processed datasets provided in GDSS~\cite{jo2022score}. These data follow to the standard pipeline~\cite{jo2022score,shi2020graphaf,luo2021graphdf}  where aromatic structures are kekulized via RDKit and hydrogen atoms are removed, treating molecules as graphs consisting solely of heavy atoms. We standardize the diffusion process by setting the number of timesteps to 1,000 for all diffusion-based models. Our model is trained using the AdamW optimizer with a constant learning rate and a weight decay of $10^{-12}$, employing EMA on model parameters to enhance stability. Following the data representation in GruM, we first encode node features using binary atom-type embeddings. Each node is represented with values in $\left[0,1\right]$, which are treated as continuous during the diffusion process. After sampling, node values are thresholded at 0.5 and subsequently converted into categorical one-hot vectors to obtain discrete atom types. For edge representation, the adjacency matrix entries are encoded as $A \in \{0,1,2,3\}$. During training, the bond values are normalized to the range $\left[0,1\right]$. After sampling, the resulting adjacency matrix is discretized and converted into a four-channel one-hot tensor that indicates the bond type of each edge.

\subsection{Visualization Settings}
For the visualization in Figure~\ref{fig:compare_path}, the initial distribution is a standard Gaussian, $x_0 \sim \mathcal{N}(0,1)$, while the target data distribution is an equally weighted mixture of two Gaussian components with means -1.5 and 1.5 and a shared standard deviation of 0.5. For both methods, we use the same diffusion schedule with $g_0=0.8$ and $g_T=0.2$. For VGD, we reduce the data-anchored coefficient from the QM9 setting of $\gamma=9$ to $\gamma=4$, since the original value leads to excessively rapid convergence and makes the intermediate dynamics difficult to visualize. For GruM, we retain its default drift coefficient $\alpha=-0.5$, which is consistently used across its experimental configurations. Both models are trained for 3,000 epochs with a batch size of 512. We use 400 temporal discretization steps and generate 5,000 samples for visualization.

\subsection{Computational Resources}
All experiments were implemented in Python 3.9.23 using PyTorch 1.12.1 and CUDA 11.6. Training and sampling were conducted on a server equipped with 3 NVIDIA 3090 GPUs, each with 24 GB of memory, an AMD EPYC 7Y43 48-Core Processor CPU, and 256 GB of system memory. Training one model required approximately 14, 57, 5, and 48 GPU hours on QM9, ZINC250k, Planar, and SBM, respectively. Developing and tuning VGD to obtain the reported dataset-specific configurations required approximately 400 GPU-hours, including exploratory runs and hyperparameter selection but excluding the repeated-seed experiments and baseline runs.

\begin{figure*}[!t]  % 注意是 figure*（带星号）
    \centering
    \includegraphics[width=0.96\textwidth]{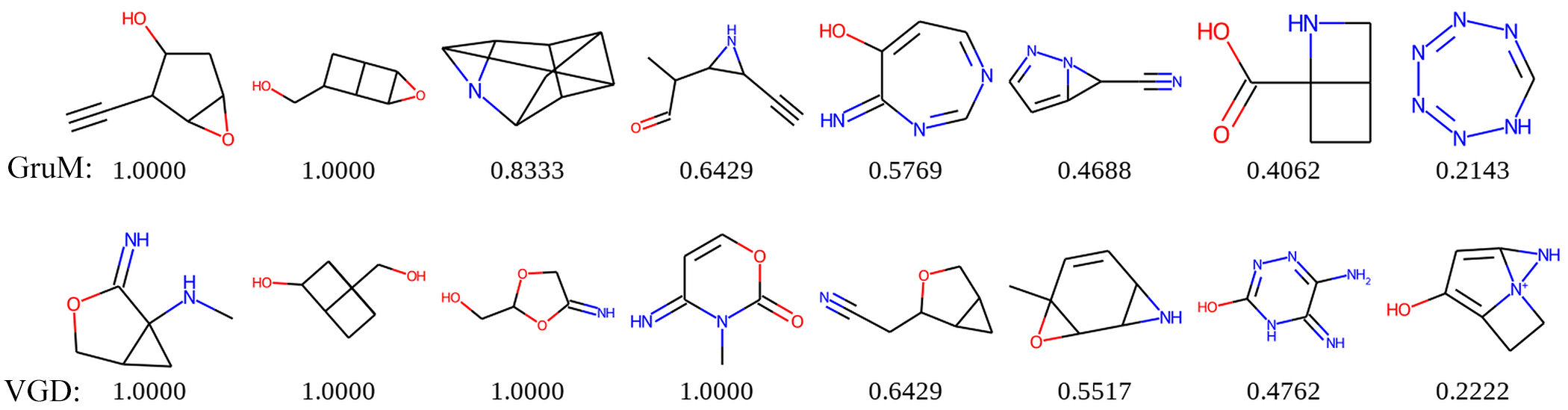}  % \textwidth = 双栏总宽度
    \caption{Generated molecules on the QM9 by the baseline (GruM, top row) and our method (VGD, bottom row), sorted in descending order of Tanimoto similarity to the training set. The values below each molecule indicate the maximum Tanimoto similarity scores.}
    \label{fig:mol_vis}
\end{figure*}
\begin{figure*}[!t]  % 注意是 figure*（带星号）
    \centering
    \includegraphics[width=0.96\textwidth]{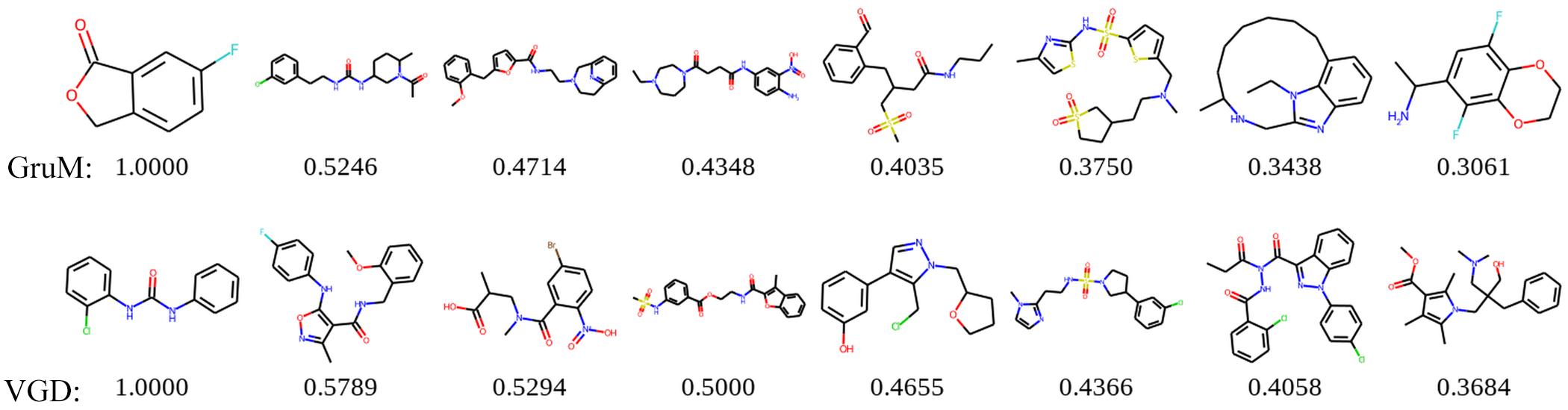}  % \textwidth = 双栏总宽度
    \caption{Generated molecules on ZINC250k by the baseline (GruM, top row) and our method (VGD, bottom row), sorted in descending order of Tanimoto similarity to the training set. The numbers below each molecule indicate the maximum Tanimoto similarity scores.}
    \label{fig:zinc_vis}
\end{figure*}
\begin{figure}[t]
    \centering
    \includegraphics[width=1\columnwidth, keepaspectratio]{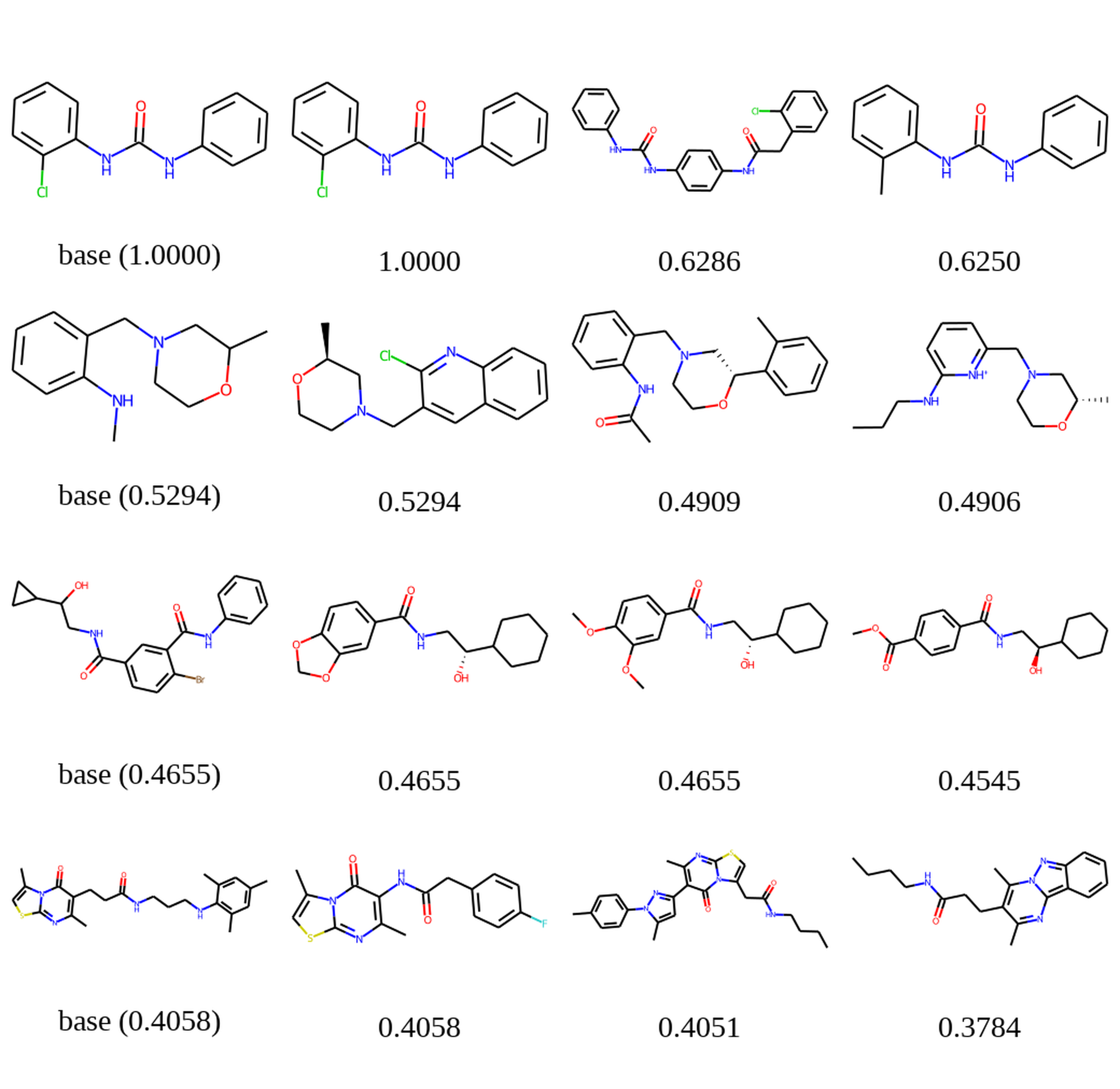}
    \caption{Visualization of generated molecules on ZINC250k and their closest training-set neighbors ranked by Tanimoto similarity.}
    \label{fig:zinc_similarity}
\end{figure}

% \subsection{Computing resources}
% The model is implemented using PyTorch. All experiments are conducted on NVIDIA GeForce RTX 3080, 3090, and 4090 GPUs.

\section{Statistical Analysis across Random Seeds}
To evaluate the robustness of the experimental results to stochastic variation in training and sampling, we repeat VGD and GruM using six matched random seeds under otherwise identical experimental settings. Table~\ref{tab:mole_statistical_results} and~\ref{tab:general_statistical_results} reports the mean and sample standard deviation of each metric. In both tables, $W$ denotes the paired Wilcoxon signed-rank statistic, and $p$ denotes the corresponding unadjusted two-sided $p$-value.

\begin{table}[t!]
\centering
\resizebox{\linewidth}{!}{
\setlength{\tabcolsep}{5pt}
\begin{tabular}{lcccc}
\toprule
Metric & GruM & VGD & $W$ & $p$ \\
\midrule
\multicolumn{5}{l}{\textbf{QM9}} \\
~~\texttt{FCD} $\downarrow$ & $0.126 \pm 0.006$ & \textbf{$0.103 \pm 0.003$} & $0.0$ & $0.03125^{*}$ \\
~~\texttt{NSPDK} $\downarrow$ & $0.0004 \pm 0.0001$ & \textbf{$0.0002 \pm 0.0000$} & $0.0$ & $0.03125^{*}$ \\
~~\texttt{Valid} $\uparrow$ & \textbf{$0.9975 \pm 0.0009$} & $0.9940 \pm 0.0005$ & $0.0$ & $0.03125^{*}$ \\
~~\texttt{Scaf.} $\uparrow$ & $0.9421 \pm 0.0052$ & \textbf{$0.9461 \pm 0.0019$} & $0.0$ & $0.03125^{*}$ \\
\midrule
\multicolumn{5}{l}{\textbf{ZINC250k}} \\
~~\texttt{FCD} $\downarrow$ & $2.192 \pm 0.030$ & \textbf{$1.213 \pm 0.013$} & $0.0$ & $0.03125^{*}$ \\
~~\texttt{NSPDK} $\downarrow$ & $0.0017 \pm 0.0001$ & \textbf{$0.0005 \pm 0.0001$} & $0.0$ & $0.03125^{*}$ \\
~~\texttt{Valid} $\uparrow$ & $0.9842 \pm 0.0013$ & \textbf{$0.9988 \pm 0.0003$} & $0.0$ & $0.03125^{*}$ \\
~~\texttt{Scaf.} $\uparrow$ & $0.5130 \pm 0.0189$ & \textbf{$0.5453 \pm 0.0124$} & $0.0$ & $0.03125^{*}$ \\
\bottomrule
\multicolumn{5}{l}{\footnotesize $^{*}p<0.05$.} \\
\end{tabular}
}
\caption{Molecular generation results on QM9 and ZINC250k across six matched-seed runs.}
\label{tab:mole_statistical_results}
\end{table}

\subsection{Molecular Graph Generation}
On QM9, VGD consistently achieves lower FCD and NSPDK values and higher scaffold similarity than GruM, demonstrating more accurate recovery of both molecular-property and graph-structural distributions. GruM retains a small advantage in validity, achieving $99.75\%$ compared with $99.40\%$ for VGD. This difference indicates that the improvements in distributional fidelity obtained by VGD are accompanied by a slight validity trade-off on this relatively small-molecule benchmark. Nevertheless, the validity of VGD remains consistently high across all runs.

The advantage of VGD is more pronounced on the more challenging ZINC250k dataset. Compared with GruM, VGD reduces the mean FCD from $2.192$ to $1.213$, corresponding to a relative reduction of approximately $44.7\%$, and reduces NSPDK from $0.0017$ to $0.0005$. VGD also improves validity by $1.46$ percentage points and increases scaffold similarity from $0.5130$ to $0.5453$. These consistent improvements indicate that VGD more accurately captures the chemical and structural distributions of larger and more diverse molecular graphs without sacrificing chemical validity.

VGD also exhibits lower or comparable standard deviations across every reported metric on both datasets. In particular, the variability of FCD, validity, and scaffold similarity is consistently smaller than that of GruM, suggesting that the performance gains are not driven by a small number of favorable runs and that the proposed diffusion dynamics provides improved run-to-run stability.

All metric comparisons yield a Wilcoxon statistic of $W=0$ and an unadjusted two-sided $p$-value of $0.03125$. For six nonzero paired observations, $W=0$ indicates complete directional consistency across the runs, although the favored method depends on the metric. In particular, the paired differences for QM9 validity, defined as VGD minus GruM, are consistently negative, showing that GruM achieves higher validity in all six runs. The identical $p$-values arise because $W=0$ corresponds to the minimum attainable exact two-sided $p$-value for six paired observations, namely $2/2^{6}=0.03125$. Therefore, the observed differences are statistically significant at the nominal $0.05$ level for each individual metric.

\subsection{General-Purpose Graph Generation}
\begin{table}[t!]
\centering
\resizebox{\linewidth}{!}{
\setlength{\tabcolsep}{5pt}
\begin{tabular}{lcccc}
\toprule
Metric & GruM & VGD & $W$ & $p$ \\
\midrule
\multicolumn{5}{l}{\textbf{Planar}} \\
~~\texttt{Deg}. $\downarrow$ & $0.0007 \pm 0.0001$ & \textbf{$0.0003 \pm 0.0001$} & $0.0$ & $0.03125^{*}$ \\
~~\texttt{Clus}. $\downarrow$ & $0.0376 \pm 0.0105$ & \textbf{$0.0347 \pm 0.0076$} & $9.0$ & $0.844$ \\
~~\texttt{Orbit} $\downarrow$ & $0.0069 \pm 0.0032$ & \textbf{$0.0044 \pm 0.0016$} & $3.0$ & $0.156$ \\
~~\texttt{Spec}. $\downarrow$ & $0.0089 \pm 0.0009$ & \textbf{$0.0072 \pm 0.0020$} & $0.0$ & $0.03125^{*}$ \\
~~\texttt{V}.\texttt{U}.\texttt{N}. $\uparrow$ & $88.67 \pm 3.63$ & \textbf{$90.00 \pm 5.92$} & $8.5$ & $0.844$ \\
\midrule
\multicolumn{5}{l}{\textbf{SBM}} \\
~~\texttt{Deg}. $\downarrow$ & $0.0013 \pm 0.0006$ & \textbf{$0.0004 \pm 0.0001$} & $0.0$ & $0.03125^{*}$ \\
~~\texttt{Clus}. $\downarrow$ & $0.0496 \pm 0.0008$ & \textbf{$0.0495 \pm 0.0013$} & $7.0$ & $0.562$ \\
~~\texttt{Orbit} $\downarrow$ & $0.0455 \pm 0.0066$ & \textbf{$0.0321 \pm 0.0140$} & $0.0$ & $0.03125^{*}$ \\
~~\texttt{Spec}. $\downarrow$ & $0.0134 \pm 0.0194$ & \textbf{$0.0043 \pm 0.0001$} & $0.0$ & $0.03125^{*}$ \\
~~\texttt{V}.\texttt{U}.\texttt{N}.$\uparrow$ & \textbf{$75.42 \pm 2.46$} & $67.92 \pm 3.68$ & $0.0$ & $0.03125^{*}$ \\
\bottomrule
\multicolumn{5}{l}{\footnotesize $^{*}p<0.05$.} \\
\end{tabular}
}
\caption{General-purpose graph generation results on Planar and SBM across six matched-seed runs.}
\label{tab:general_statistical_results}
\end{table}

To evaluate the robustness of VGD on general-purpose graph generation, we conduct six repeated runs with matched random seeds on the Planar and SBM datasets. Table~\ref{tab:general_statistical_results} reports the mean and sample standard deviation of each metric, together with two-sided paired Wilcoxon signed-rank tests over the seed-wise results.

On Planar, VGD achieves better mean performance than GruM across all five metrics. In particular, VGD reduces the mean degree discrepancy from $0.0007$ to $0.0003$ and the spectral discrepancy from $0.0089$ to $0.0072$. Both improvements are consistent across the paired runs and reach nominal statistical significance, with $W=0$ and $p=0.03125$. VGD also obtains lower mean clustering and orbit discrepancies, decreasing Clus. from $0.0376$ to $0.0347$ and Orbit from $0.0069$ to $0.0044$. However, the corresponding differences are not statistically significant, indicating that their magnitudes vary across random seeds. Similarly, VGD attains a slightly higher mean V.U.N. score of $0.9000$, compared with $0.8867$ for GruM, but this difference is not statistically significant. Overall, the repeated runs provide the strongest evidence for improvements in degree-distribution and spectral fidelity, while the results for clustering, orbit statistics, and V.U.N. should be interpreted as favorable average trends rather than statistically reliable advantages.

On SBM, VGD demonstrates clear advantages in the structural-distribution metrics. It reduces the mean degree discrepancy from $0.0013$ to $0.0004$, the orbit discrepancy from $0.0455$ to $0.0321$, and the spectral discrepancy from $0.0134$ to $0.0043$. All three differences yield $W=0$ and $p=0.03125$, showing that VGD consistently outperforms GruM across the matched runs on these metrics. The particularly small standard deviation of the spectral score for VGD also indicates stable recovery of global graph structure, whereas the substantially larger variation observed for GruM suggests greater sensitivity to random initialization. By contrast, the difference in clustering discrepancy is negligible and not statistically significant, with mean values of $0.0495$ and $0.0496$ for VGD and GruM, respectively.

The SBM results also reveal a trade-off between structural fidelity and the composite V.U.N. metric. Although VGD more accurately matches the reference degree, orbit, and spectral distributions, GruM achieves a higher mean V.U.N. score of $0.7542$, compared with $0.6792$ for VGD. The paired Wilcoxon test indicates that this difference is nominally significant. Therefore, the improved distributional alignment of VGD on SBM does not translate into a higher proportion of samples that simultaneously satisfy validity, uniqueness, and novelty. This result suggests that VGD prioritizes accurate recovery of the underlying community and global structural statistics, while GruM retains an advantage in the composite sample-level diversity criterion on this dataset.

Taken together, the repeated experiments confirm that VGD provides consistent improvements in several complementary measures of structural fidelity. The evidence is particularly strong for degree and spectral distributions on both datasets, as well as orbit statistics on SBM. Nevertheless, the nonsignificant results for several Planar metrics and the lower V.U.N. score on SBM indicate that the advantage of VGD is primarily reflected in distributional and topological accuracy rather than uniformly across every generation criterion.

\section{Visualization of Generated Molecules}\label{sec:mol_vis}
To illustrate the structural diversity and similarity of molecules generated by our model (VGD) and the baseline (GruM), we visualize representative generated molecules, with their maximum Tanimoto similarity to the training set labeled alongside each structure.

Figure~\ref{fig:mol_vis} visualizes eight molecular graphs generated by each method (GruM on the top row and VGD on the bottom row). These molecular graphs are sampled from the 10,000 generated samples of each model, specifically selected from the last 5,000 samples using identical indices, and sorted in descending order of their Tanimoto similarity. The value shown below each molecular graph denotes its maximum Tanimoto similarity to any training molecule, where a score of 1.0 indicates an identical match. As illustrated in the figure, VGD consistently maintains higher similarity across a wider range of molecules and demonstrates more balanced and robust generation performance compared to GruM.

Figure~\ref{fig:zinc_vis} visualizes eight molecular structures generated by GruM (top) and VGD (bottom) on ZINC250k dataset, which is a benchmark dataset focusing on drug-like molecules with diverse and complex scaffolds. The molecules are sampled from 10,000 generated samples by selecting one every 1250 samples, and are sorted in descending order according to the maximum Tanimoto similarity to the training set. As shown in the figure, our method consistently produces molecules with higher similarity compared to GruM, while simultaneously generating structures with complex and diverse scaffolds.

Figure~\ref{fig:zinc_similarity} presents a similarity visualization on the ZINC-250K dataset, where representative generated molecules are selected from a sorted list based on their maximum Tanimoto similarity to the training set, sampling one molecule every 2,500 entries. For each selected ``base'' molecule, we display the top three nearest neighbors in the training set according to Morgan fingerprint Tanimoto scores. As the similarity decreases, the structural variation increases while preserving valid chemical functional groups, indicating that the model successfully balances novelty and structural realism. This visualization confirms that the generated distribution is both diverse and aligned with the chemical space.
\section{Visualization of General Graph}\label{sec:gen_vis}
\begin{figure}[!t] 
    \centering
    \includegraphics[width=0.96\linewidth]{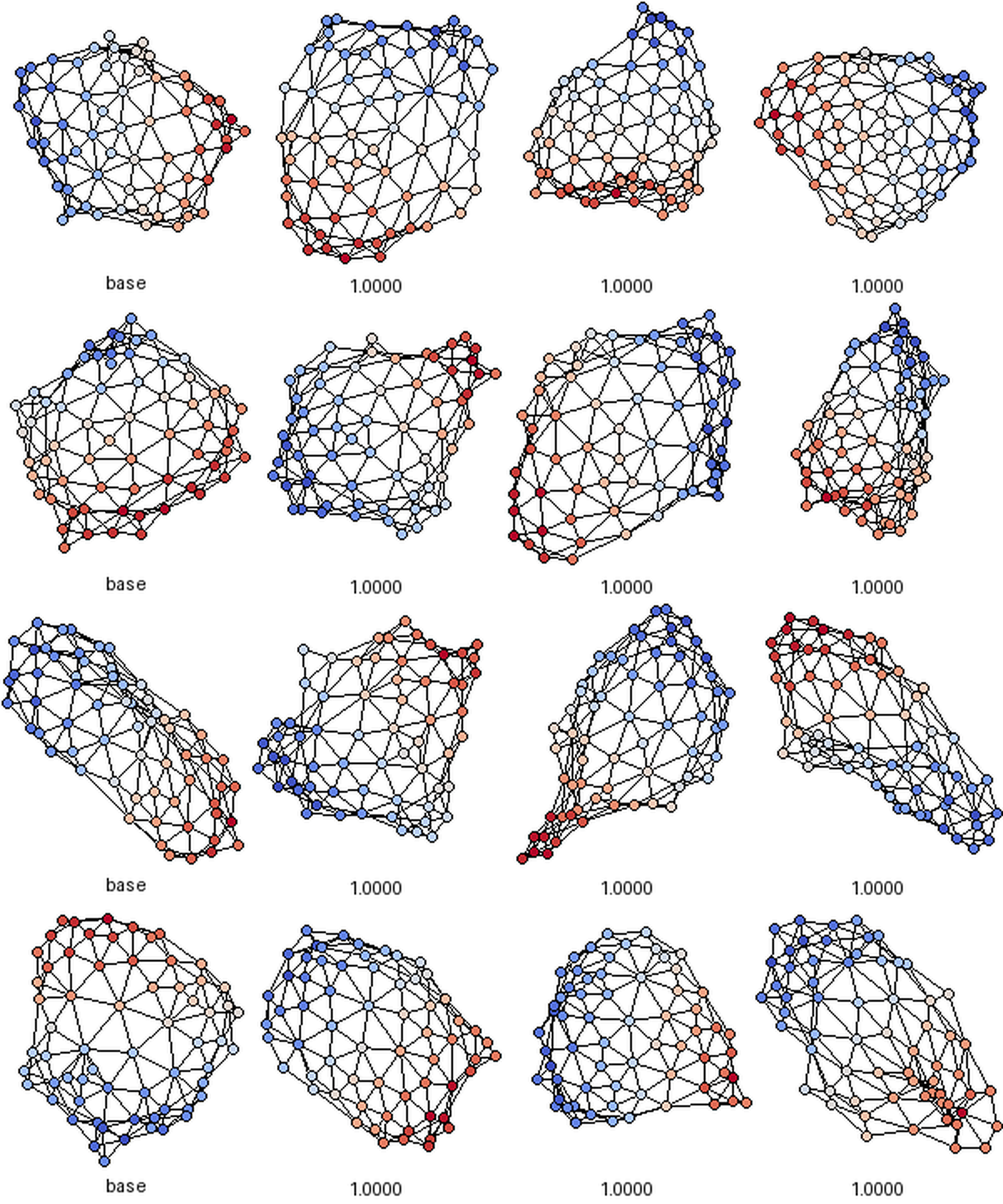}  % \textwidth = 双栏总宽度
    \caption{
Visualization of generated graphs and their most similar training samples on Planar. Each row displays one generated graph (left) together with its top-3 closest training graphs (right), ranked by Tanimoto similarity.
    }
    \label{fig:planar_vis}
\end{figure}
Figure~\ref{fig:planar_vis} visualizes the structural correspondence between the generated graphs on Planar (leftmost column, labeled \textit{``base''}) and their top-3 most similar training graphs. The value shown below each training graph denotes its Tanimoto similarity to the generated graph. The visualization highlights two key observations:
\begin{itemize}[leftmargin=*]
    \item \textbf{Strict adherence to planar constraints.} The generated graphs fully preserve the defining planar property of having \emph{non-crossing edges}, and their node connectivity patterns are consistent with those observed in the training samples. This confirms that the model accurately captures the fundamental structural rules required for planar graph generation. The apparent edge crossings in Figure~\ref{fig:planar_vis} arise from the force-directed layouts used solely for visualization and do not indicate violations of planarity.
    \item \textbf{Perfect structural consistency despite visual layout differences.} In all cases, the Tanimoto similarity reaches \textbf{1.0000}, indicating an exact match in topological structure between generated and retrieved training graphs. Notably, although the graph layouts appear visually different due to the use of independent layout algorithms, their underlying connectivity patterns are identical. This demonstrates that the model learns the true structural distribution of planar graphs rather than simply replicating visual appearances.
\end{itemize}
\begin{figure}[!t]  % 注意是 figure*（带星号）
    \centering
    \includegraphics[width=0.96\linewidth]{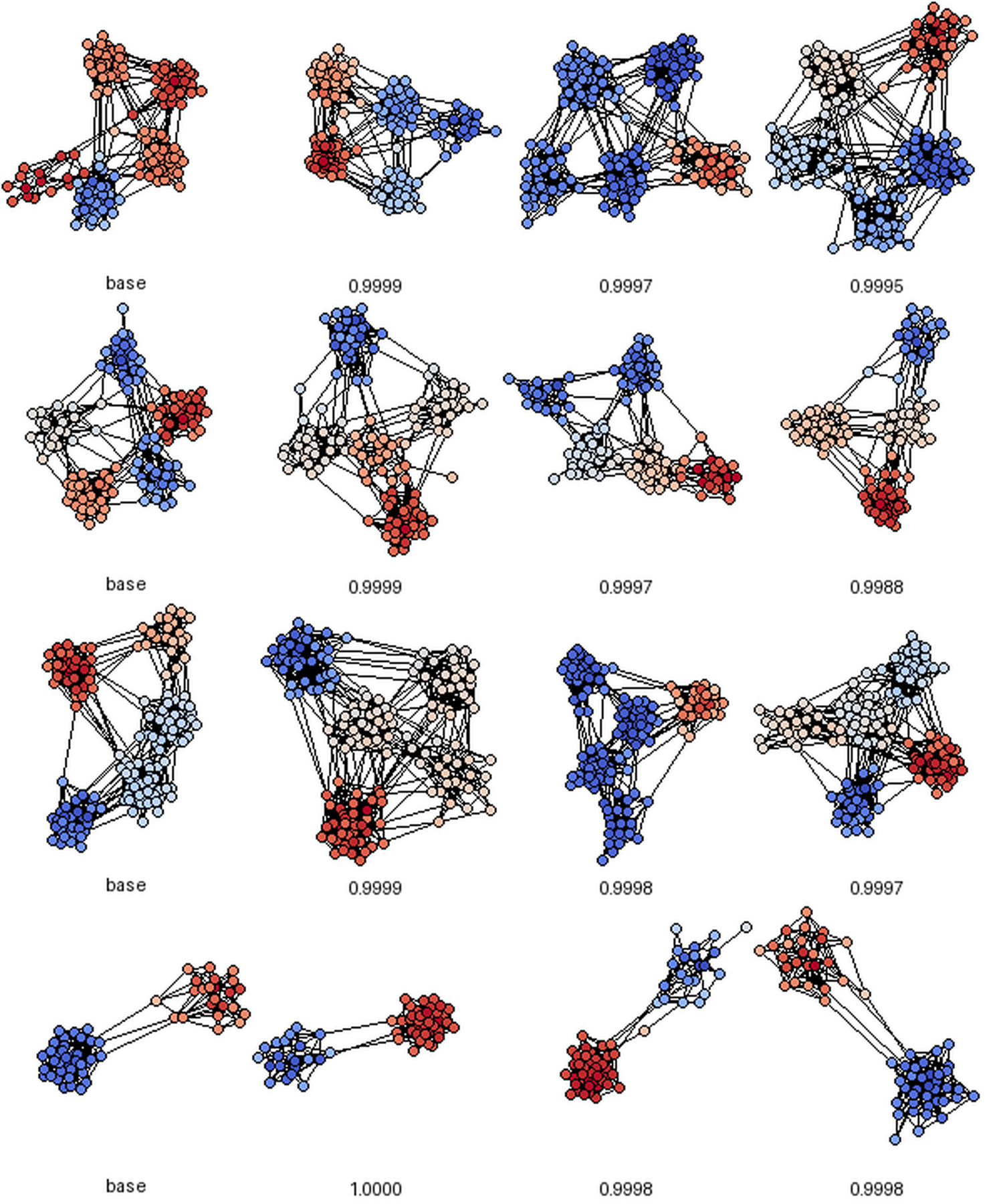}  % \textwidth = 双栏总宽度
    \caption{
Visualization of generated graphs and their most similar training samples on SBM. Each row shows one generated graph (left) and its top-3 closest training graphs ranked by Tanimoto similarity (right).
}\label{fig:sbm_vis}
\end{figure}

Figure~\ref{fig:sbm_vis} visualizes the structural alignment between the generated graphs on SBM (leftmost column, labeled \textit{``base''}) and their top-3 most similar training graphs. The value shown below each training graph denotes its Tanimoto similarity to the generated graph, where 1.0000 indicates a perfect structural match. The visualization demonstrates that:
\begin{itemize} [leftmargin=*]
    \item \textbf{High structural consistency.} The generated graphs exhibit clear community structures, where different colors correspond to different communities. The node coloring and connectivity patterns collectively reflect strong community cohesion and closely match those of the training graphs, confirming that the model effectively captures the core community patterns of SBM.
    \item \textbf{Controllable diversity.} While maintaining strong structural similarity, the generated graphs are not identical to the training samples. This indicates a balanced trade-off in community-based graph generation between fitting the training distribution and avoiding redundant replication.
\end{itemize}

\end{document}